\documentclass[11pt, abstract=on]{article}

\usepackage{authblk} 

\usepackage[ngerman, english]{babel} 
\usepackage[utf8]{inputenc} 
\usepackage[T1]{fontenc}
\usepackage{comment}

\usepackage{lmodern} 

\usepackage[pdfpagelabels, hidelinks]{hyperref} 
\usepackage{enumerate} 

\usepackage{amsmath, amsthm, amssymb,amsfonts} 
\usepackage{dsfont} 
\usepackage{mathtools}

\newcommand\blfootnote[1]{
	\begingroup
	\renewcommand\thefootnote{}\footnote{#1}
	\addtocounter{footnote}{-1}
	\endgroup
}

\usepackage{xcolor} 
\definecolor{seeblau}{RGB}{89,199,235} 
\definecolor{Seeblau}{RGB}{0,169,224} 
\definecolor{seeb}{RGB}{166,225,244} 

\usepackage{orcidlink}

\theoremstyle{plain} 
\newtheorem{thm}{Theorem}[section]
\newtheorem{lem}[thm]{Lemma}
\newtheorem{prop}[thm]{Proposition}

\newtheorem*{claim}{Claim}

\newtheorem*{FundThm}{Fundamental Theorem of Statistical Learning}

\theoremstyle{definition}
\newtheorem{defn}[thm]{Definition}

\newtheorem{exa}[thm]{Example}
\AtEndEnvironment{exa}{\hfill\ensuremath{\diamond}}
\newtheorem{rem}[thm]{Remark}
\AtEndEnvironment{rem}{\hfill\ensuremath{\diamond}}

\theoremstyle{definition} 
\newtheorem{frag}{Question}[section]

\numberwithin{equation}{section}

\renewcommand{\restriction}{\mathord{\upharpoonright}}

\let\phi\varphi

\newcommand\DD{\mathbb{D}}
\newcommand\EE{\mathbb{E}}
\newcommand\NN{\mathbb{N}}

\newcommand\RR{\mathbb{R}}

\newcommand\PP{\mathbb{P}} 

\newcommand\A{\mathcal{A}}
\newcommand\B{\mathcal{B}}

\newcommand\DDD{\mathcal{D}}

\newcommand\HH{\mathcal{H}}

\newcommand{\LL}{\mathcal{L}}
\newcommand\M{\mathcal{M}}

\newcommand\Pot{\mathcal{P}}
\newcommand\R{\mathcal{R}}
\newcommand\U{\mathcal{U}}
\newcommand\X{\mathcal{X}}

\newcommand\Z{\mathcal{Z}}

\newcommand{\Lor}[0]{\mathcal{L}_{\mathrm{or}}}
\newcommand{\Lr}[0]{\mathcal{L}_{\mathrm{r}}}

\DeclareMathOperator{\er}{er}

\DeclareMathOperator{\opt}{opt}

\DeclareMathOperator{\fin}{fin}

\newcommand{\1}{\mathds{1}}

\newcommand{\ul}[1]{\underline{#1}}

\newcommand{\brackets}[1]{\!\left(#1\right)} 

\title{\Large\textbf{{Measurability in the Fundamental Theorem\\ of Statistical Learning}} \\ {\large(with an appendix by Laura Wirth)}}

\author[ii,i]{Lothar Sebastian Krapp \orcidlink{0000-0003-3102-1923}\,}
\author[i]{Laura Wirth \orcidlink{0000-0003-2871-5676}\,}
\affil[i]{\,Fachbereich Mathematik und Statistik, Universität Konstanz, Germany}
\affil[ii]{\,Institut für Interdisziplinäre Sprachevolutionswissenschaft, Universität Zürich, Switzerland}

\date{}

\usepackage[skip=8pt+2pt-2pt]{parskip}

\begin{document}
	
	\pagenumbering{arabic}
	
	
	\maketitle
	\blfootnote{Math Subject Classification (2020): Primary 68T05, 03C64; Secondary 28A05, 28A20, 03C98, 68T27, 12J15. Key Words: PAC Learning, VC Dimension, Uniform Convergence, Measurability, NIP, o-Minimality.
	}
	\blfootnote{Corresponding Author: Laura Wirth, laura@uni-konstanz.de.
	}
	\vspace{-1.3cm}
	
	\renewcommand\abstractname{Abstract}
	\begin{abstract}
		
		\noindent
		\small{The Fundamental Theorem of Statistical Learning states that a hypothesis space is PAC learnable if and only if its VC dimension is finite. For the agnostic model of PAC learning, the literature so far presents proofs of this theorem that often tacitly impose several measurability assumptions on the involved sets and functions. We scrutinize these proofs from a measure-theoretic perspective in order to explicitly extract the assumptions needed for a rigorous argument. This leads to a sound statement as well as a detailed and self-contained proof of the Fundamental Theorem of Statistical Learning in the agnostic setting, showcasing the minimal measurability requirements needed. As the Fundamental Theorem of Statistical Learning underpins a wide range of further theoretical developments, our results are of foundational importance: A careful analysis of measurability aspects is essential, especially when the theorem is used in settings where measure-theoretic subtleties play a role. We particularly discuss applications in Model Theory, considering NIP and o-minimal structures. Our main theorem presents sufficient conditions for the PAC learnability of hypothesis spaces defined over o-minimal expansions of the reals. This class of hypothesis spaces covers all artificial neural networks for binary classification that use commonly employed activation functions like ReLU and the sigmoid function.}
	\end{abstract}
	
	\pagebreak

	
	\tableofcontents
	
	\section{Introduction}
	
	In 1984, Valiant~\cite{Valiant} introduced a formal model of learning based on randomly generated examples, which is referred to as \emph{Probably Approximately Correct (PAC) Learning}. Conceptually, a class of (indicator) functions is PAC learnable if for any finitely many randomly generated examples there exists a suitable generalization function within the class. Blumer, Ehrenfeucht, Haussler and Warmuth~\cite{Blumer} established in 1989 that an indicator function class is PAC learnable if and only if its VC dimension is finite. Their result is commonly referred to as the \emph{Fundamental Theorem of Statistical Learning}. Both the argument in \cite{Blumer} and all following proofs of this theorem in the literature rely on ``certain broadly applicable measurability conditions'' (Ackerman, Asilis, Di, Freer and Tristan~\cite[page~1]{AADFT}). The VC dimension -- introduced by Vapnik and Chervonenkis\footnote{Due to different transliteration from the Cyrillic, various spellings of this name are common in Latin script.}~\cite{VC1968} in 1968 -- is a parameter for measuring the combinatorial size of an indicator function class. Laying the basis for the Fundamental Theorem of Statistical Learning, Vapnik and Chervonenkis~\cite{VC1971} proved in 1971 that finiteness of the VC dimension is a necessary and sufficient condition for uniform convergence of empirical means. In 1992, Laskowski~\cite{Laskowski} revealed a striking analogy between NIP formulas and classes having finite VC dimension. Here, NIP stands for `\emph{not the independence property}' (for details see Poizat~\cite[\S\,12.4]{Poizat}). This triangular relationship of the notions NIP, VC dimension and PAC learnability inspired a fruitful exchange between Model Theory and Statistical Learning Theory, as ``connections between the VC theory and NIP have developed extensively with important notions from VC-theory adapted to the model-theoretic setting and vice versa'' (Chase and Freitag~\cite[page~323]{Chase-Freitag}). For instance, Steinhorn~\cite[page~27]{Steinhorn} points out that ``a definable family of definable sets $\mathcal{C}$ in an o-minimal structure is PAC learnable''. For drawing such connections special care has to be taken, as the Fundamental Theorem of Statistical Learning applies only to classes satisfying certain measurability conditions. In \cite{Blumer}, a precise analysis of the measurability requirements is provided. The authors introduce the notion of \emph{well-behaved} classes\footnote{This is a measure-theoretic property that is developed in \cite[Appendix~A1]{Blumer}. We introduce a similar property in Section~\ref{sec::fundthm}.}, to which the Fundamental Theorem of Statistical Learning applies. Although in \cite[page~935]{Blumer} this property is described as ``benign'', it requires ``the measurability of complex sets of samples'' (Ben-David, Benedek and Mansour~\cite[page~13]{Ben-David-Benedek-Mansour}). Thus, the well-behavedness assumption constitutes a ``complicated measurability constraint'' (Ben-David, Itai and Kushilevitz~\cite[page~248]{Ben-David-Itai-Kushilevitz}).
	
	The underlying PAC learning model considered in \cite[\S\,2]{Blumer} relies on deterministic labels. More precisely, the samples on which the learning is based are generated according to distributions on the instance space and are equipped with labels that are generated according to a target function. In contrast, following more recent literature such as Shalev-Shwartz and Ben-David~\cite[\S\,3.2.1]{UnderstandingML}, we work with a model of PAC learning that is \emph{agnostic} in various respects. This model omits target functions and is instead based on working with an arbitrary set of \emph{joint} distributions on the sample space. Thus, the labels are not deterministic but instead also generated randomly. The agnostic PAC model therefore provides a more general and realistic description of formal learning (cf.\ Vidyasagar~\cite[\S\,3.3.1]{Vidyasagar}). In particular, it is able to deal with noisy and corrupted data (cf.\ von Luxburg and Schölkopf~\cite[page~654\;f.]{vonLuxburg-Schoelkopf}). For further details see Remark~\ref{rem::joint-distributions}, cf.\ also \cite[Appendix~A.3]{Blumer}.
	
	Several expositions in Statistical Learning acknowledge that measurability subtleties demand a certain amount of attention (see e.g.\ Haussler~\cite[page~190]{Haussler1988}, Mendelson and Smola~\cite[page~5]{Mendelson-Smola}, Pestov~\cite[page~1141]{Pestov} and Vidyasagar~\cite[page~56]{Vidyasagar}). Some authors assume that the function class under consideration is \emph{permissible}\footnote{This is a measure-theoretic property that is developed in Pollard~\cite[Appendix~C]{Pollard}.} (e.g. Haussler~\cite[\S\,9.2]{Haussler1992} and Lee~\cite[\S\,2.3]{Lee1996}) in order to prove according results. From a purely theoretical point of view, this assumption confines consideration to function classes with a rather specific structure. In practice, most function classes arising in applications satisfy this condition. 
	
	Our initial motivation for this work was to explore theoretical applications of the Fundamental Theorem of Statistical Learning, particularly in the model-theoretic context of linearly ordered structures. However, the available literature so far lacks a systematic analysis of the \emph{minimal} measurability requirements of the Fundamental Theorem of Statistical Learning for \emph{agnostic} PAC learning. In this article, we fill this gap by identifying these requirements, and thereby describing the most general measure-theoretic framework in which the result and its proof remain valid. More precisely, we present a self-contained proof that addresses the relevant measurability issues in a detailed and precise manner. In doing so, we build a foundation upon which future learnability results in the context of Model Theory can be precisely formulated.
	
	Throughout this paper, we concentrate on binary classification problems, as these are well-studied within Supervised Learning, a subbranch of Machine Learning. In this context, (supervised) learning refers to the process of inferring a general rule on how to classify certain objects by observing examples. Our focus lies on establishing qualitative learnability results, and we therefore do not address quantitative aspects such as explicit sample complexity bounds.
	
	Our paper is structured as follows. In Section~\ref{sec::fram-prelim} we describe our learning framework and gather basic preliminaries from Statistical Learning Theory. In particular, we introduce the notions that are fundamental for the results we establish in Section~\ref{sec::fundthm}: (agnostic) PAC Learning, Uniform Convergence Property and VC Dimension. Special emphasis is put on measure-theoretic aspects of these concepts. In Section~\ref{sec::fundthm} we present a sound statement as well as a detailed and self-contained proof of the Fundamental Theorem of Statistical Learning (Theorem~\ref{thm::fundthm-conditions}). Its result relies on some measurability assumptions that are described in Definition~\ref{defn::well-behaved}. The proof comprises the verification of several implications (see Theorem~\ref{thm::VC-UCP}, Theorem~\ref{thm::UCP-NMSE}, Proposition~\ref{prop::NMSE-existence} and Theorem~\ref{thm::PAC-VC}). A key step is to prove that any hypothesis space with finite VC dimension has the uniform convergence property, provided certain measurability requirements are met (see Theorem~\ref{thm::VC-UCP-help} and Theorem~\ref{thm::VC-UCP}). This is the part of the argument that demands most of the technical work and is precisely where measurability assumptions are crucial. Section~\ref{sec::model-theory} initiates an examination of applications of the Fundamental Theorem of Statistical Learning in Model Theory. We introduce \emph{definable} hypothesis spaces over first-order structures (Definition~\ref{defn::dfblhc}), which always have finite VC dimension if the underlying structure has NIP (Proposition~\ref{prop::nipvcmodeltheory}). Our main theorem (Theorem~\ref{thm::o-minimal-PAC}) establishes sufficient conditions for the PAC learnability of hypothesis spaces defined over o-minimal expansions of the reals. This class of hypothesis spaces encompasses artificial neural networks whose activation functions are definable over the real exponential field (Example~\ref{exa::rexp}).We complete this work by gathering open questions in Section~\ref{sec::further-work}.  
	
	\textbf{CRediT Authorship Contribution Statement:} Lothar Sebastian\linebreak Krapp: conceptualization (supporting); funding acquisition (lead); investigation (supporting); project administration (lead); supervision (lead); writing – original draft (supporting), writing – review \& editing (equal). Laura Wirth: conceptualization (lead); investigation (lead); writing – original draft (lead), writing – review \& editing (equal).
	
	\textbf{Funding:} This work is part of the research project \emph{Fundamentale Grenzen von Lernprozessen in künstlichen neuronalen Netzen}, funded by Vector Stiftung within the program \emph{MINT-Innovationen 2022}.
	
	\textbf{Acknowledgements:} We thank Matthias C.~Caro, Salma Kuhlmann and\linebreak Tobias~Sutter for several helpful discussions and for valuable advice on relevant literature. Moreover, we thank anonymous referees for providing several helpful comments and references. The second author would also like to extend special thanks to Salma Kuhlmann for the supervision of her doctoral project, which this work is part of. Additionally, the second author is grateful to David Pollard, Itay Kaplan, Markus Kunze, Tim Seitz and Patrick Michalski for pointing out various challenges related to measurability, clarifying some subtleties, and suggesting potential solutions during personal exchanges.
	
	\textbf{Conflict of Interest:} The authors declare no conflict of interest. The funders had no role in the design and conduct of the study; preparation, review, or approval of the manuscript; and decision to submit the manuscript for publication.
	
	\section{Framework and Preliminaries}\label{sec::fram-prelim}
	
	The notions from Statistical Learning Theory that we introduce in this section are based on the expositions in Anthony and Bartlett~\cite{Anthony-Bartlett}, Shalev-Shwartz and Ben-David~\cite{UnderstandingML} as well as Vidyasagar~\cite{Vidyasagar}. 
	
	We denote by $\NN$ the set of positive natural numbers and we set $\NN_0=\NN\,\dot{\cup}\,\{0\}$. Given $m\in\NN$, we denote by $[m]$ and $[m]_0$ the sets $\{1,\dots,m\}$ and $\{0,\dots,m\}$, respectively. Often we would like to emphasize that some object is a tuple and we then use underlined letters such as $\ul{z}$. The power set of a set $A$ is denoted by $\Pot(A)$, and the set of finite subsets of $A$ by $\Pot_{\fin}(A)$. We denote by $f\restriction_A$ the restriction of a function $f$ to a subset $A$ of its domain. The indicator function $\1_A$ of a set $A$ maps any element of $A$ to $1$ and everything else to $0$. Given a measurable space $(\Omega,\Sigma)$ and an element $\omega\in\Omega$, we denote by $\delta_\omega$ the Dirac measure defined by
	$$\delta_\omega\colon \Sigma\to\{0,1\}, A\mapsto \delta_\omega(A)=\1_A(\omega).$$
	Given a probability space $(\Omega,\Sigma,\PP)$, a random variable is a $\Sigma$--measurable\footnote{A map $X\colon\Omega\to\RR$ is $\Sigma$--measurable if $X^{-1}(B)\in\Sigma$ for any Borel set $B\subseteq\RR$.} map $X\colon\Omega\to\RR$, and we denote the expected value of $X$ by $\EE_{\omega\sim\PP}[X(\omega)]$, if it exists. Since we work in a measure-theoretic context, we point out that our results are entirely established within the set-theoretic framework of ZFC. 
	
	Throughout this section, we consider a \emph{learning problem}, which is specified by the tuple $(\X,\Sigma_\Z,\DDD,\HH)$, where $\X$ is a non-empty set called \textbf{instance space}, $\Sigma_\Z$ is a $\sigma$--algebra on the \textbf{sample space} $\Z=\Z_\X=\X\times\{0,1\}$ fulfilling $\Pot_{\fin}(\Z)\subseteq\Sigma_\Z$, $\DDD$ is a subset of the set $\DDD^\ast$ containing all (probability) distributions defined on the measurable space $(\Z,\Sigma_\Z)$. The \textbf{hypothesis space} $\HH$ is a non-empty subset of the set $\{0,1\}^\X$ of functions from $\X$ to $\{0,1\}$, and its elements $h\in\HH$ are referred to as \textbf{hypotheses}.
	
	\begin{rem}\label{rem::basic-measurability-assumptions}\
		\begin{enumerate}[(a)]
			\item The instance space $\X$ contains the objects to be classified into positive and negative examples using the binary \emph{labels} $0$ and $1$. 
			
			\item We assume that $\Pot_{\fin}(\Z)\subseteq\Sigma_\Z$, or equivalently $\{z\}\in\Sigma_\Z$ for any $z\in\Z$. This is true if $\Z$ is a Hausdorff topological space -- particularly, if $\Z$ is a metric space -- and $\Sigma_\Z$ is the corresponding Borel $\sigma$--algebra\footnote{Given a topological space $(\Z,\tau)$, the corresponding Borel $\sigma$--algebra is the smallest $\sigma$--algebra containing $\tau$ as a subset.}. 
			Note further that our assumption implies $\Sigma_\Z=\Pot(\Z)$ if $\X$ is countable. In the literature it is common to assume that $\Z$ is a metric space and that $\Sigma_\Z$ is the according Borel $\sigma$--algebra (see Haussler~\cite[page~91]{Haussler1992}).
			
			\item The distributions in $\DDD$ represent the possible ``states of nature'' that might underlie the generation of the samples on which the learning is based (Haussler~\cite[page~93]{Haussler1992}). We can always choose to work with the set $\DDD^\ast$ containing \emph{all} distributions to obtain distribution-free results. 
		\end{enumerate}
		\vspace{-0.7em} 
	\end{rem}
	
	For $h\in\HH$ we denote by $\Gamma(h)=\{(x,y)\in\Z\mid h(x)=y\}$ the graph of $h$.
	
	While in later sections we will specifically point out measurability requirements on the hypotheses, \textbf{throughout this section, we assume that for any hypothesis $h\in\HH$ we have $\Gamma(h)\in\Sigma_\Z$.} 
	
	In our context, \emph{learning} is based on processing finitely many \emph{samples}
	$$(x_1,y_1),\dots,(x_m,y_m)\in\Z$$
	that are drawn independently at random according to some probability distribution $\DD\in\DDD$. Thus, the \emph{multi-sample} $\ul{z}=((x_1,y_1),\dots,(x_m,y_m))\in\Z^m$ is drawn according to $\DD^m$, where $\DD^m$ denotes the product measure
	$\bigotimes_{i=1}^m\DD,$
	which is defined as usual on the product $\sigma$--algebra\footnote{The product $\sigma$--algebra $\Sigma_\Z^m=\bigotimes_{i=1}^m\Sigma_\Z$ on $\Z^m$ is the smallest $\sigma$--algebra on $\Z^m$ containing all sets of the form $\bigtimes\limits_{i\in[m]}\! C_i$, where $C_i\in\Sigma_\Z$ for $i\in[m]$.} $\Sigma_\Z^m$. 
	
	\begin{rem}\label{rem::joint-distributions}
		As we consider \emph{joint} distributions $\DD$ on the sample space $\Z$, it is possibly the case that for an instance $x\in\X$ both $(x,0)$ and $(x,1)$ have positive probability. In other words, the labels of instances are not deterministic or provided by a specific target function but are random themselves. This also allows us to deal with ``noisy labels or corrupted data''. The underlying learning paradigm is sometimes referred to as \emph{agnostic} (see \cite[\S\,3.2.1]{UnderstandingML}) or \emph{model-free} (see \cite[\S\,3.3]{Vidyasagar}, respectively). In contrast, Blumer, Ehrenfeucht, Haussler and Warmuth~\cite[\S\,2]{Blumer} work with a learning model that only involves distributions on the instance space $\X$. Thus, the instances $x_1,\dots,x_m\in\X$ are drawn independently at random according to some probability distribution $\PP$ on a $\sigma$--algebra on $\X$. The labels $y_1,\dots,y_m$ are then generated according to a fixed target function $t\in\{0,1\}^\X$. 
	\end{rem}
	
	The goal of our learning paradigm is to predict a hypothesis in $\HH$ providing a general rule on how to classify the instances in $\X$ based on a given multi-sample. Formally, this generalization scheme is captured by a \textbf{learning function} for $\HH$, which is any map of the form
	$$\A\colon\bigcup\limits_{m\in\NN}\Z^m\to\HH.$$
	We wish to find a learning function that reflects a ``reasonable'' generalization of the information provided by multi-samples.\footnote{In the literature, such a map $\A$ is also referred to as ``learner'' (see \cite{Anthony-Bartlett}) or ``(learning) algorithm'' (see \cite{UnderstandingML} and \cite{Vidyasagar}), among other terms. Note that we do not place any restrictions on the nature of the function $\A$ such as e.g.\ computability.} 
	
	The \emph{quality} of the determined hypotheses is measured by their \emph{error}, also known as \emph{true error}, \emph{generalization error} or \emph{risk}. Given $\DD\in\DDD^\ast$ and $h\in\HH$, the \textbf{(true) error} of $h$ according to $\DD$ is defined by 
	$$\er_\DD(h):=\DD(\Z\setminus\Gamma(h))=\DD(\{(x,y)\in\Z\mid h(x)\neq y\}),$$
	i.e.\ the probability of \emph{misclassification}. The map $\er_\DD\colon\HH\to[0,1]$ is well-defined, since we assume that $\Gamma(h)\in\Sigma_\Z$ for any $h\in\HH$. While $\HH$ does not necessarily contain a hypothesis $h$ with $\er_\DD(h)=0$, one can aim at finding a hypothesis whose error is close to the ``best (possible) performance'' value
	$$\opt_\DD(\HH)=\inf\limits_{h\in\HH}\er_\DD(h),$$
	which is referred to as \textbf{approximation error} of $\HH$ (with respect to $\DD$). 
	
	Usually, the underlying distribution $\DD$ is unknown, and hence the true error is not directly available. However, a useful estimate of the true error that can always be calculated is the \emph{sample error}, also known as \emph{training error}, \emph{empirical error} or \emph{empirical risk}. Given $m\in\NN$, $\ul{z}=(z_1,\dots,z_m)\in\Z^m$ and $h\in\HH$, the \textbf{sample error} of $h$ on $\ul{z}$ is given by 
	$$\hat{\er}_{\ul{z}}(h):=\frac{1}{m}\sum\limits_{i=1}^m \ell(h,z_i),$$
	where $\ell\colon\HH\times\Z\to\{0,1\}$, referred to as \textbf{loss function}, is defined by 
	$$\ell(h,z)=\1_{\Z\setminus\Gamma(h)}(z)=\begin{cases}
		1&\text{if }h(x)\neq y, \\
		0&\text{otherwise,}
	\end{cases}$$
	for $h\in\HH$ and $z=(x,y)\in\Z$. We denote the minimal sample error of $\HH$ by
	$$\hat{\opt}_{\ul{z}}(\HH)=\inf\limits_{h\in\HH}\hat{\er}_{\ul{z}}(h).$$
	As $\hat{\er}_{\ul{z}}(h)$ can only take on finitely many values, we observe that $$\hat{\opt}_{\ul{z}}(\HH)=\inf\limits_{h\in\HH}\hat{\er}_{\ul{z}}(h)=\min\limits_{h\in\HH}\hat{\er}_{\ul{z}}(h).$$
	
	\begin{rem}\label{rem::sample-error-measurable}
		Let $h\in\HH$. The map $\ell(h,\cdot)$ is $\Sigma_\Z$--measurable, as $$\ell(h,\cdot)^{-1}(\{0\})=\1_{\Z\setminus\Gamma(h)}^{-1}(\{0\})=\Gamma(h)\in\Sigma_\Z,$$
		and we compute 
		$$\EE_{z\sim\DD}[\ell(h,z)]=\EE_{z\sim\DD}[\1_{\Z\setminus\Gamma(h)}(z)]=\DD(\Z\setminus\Gamma(h))=\er_\DD(h).$$
		Furthermore, the map 
		$$\hat{\er}_\square(h)\colon\Z^m\to\left\{\left.\tfrac{k}{m}\ \right|\, k\in[m]_0\right\},\ \ul{z}\mapsto \hat{\er}_{\ul{z}}(h)$$ 
		is $\Sigma_\Z^m$--measurable. In fact, for any $k\in[m]_0$ we compute
		$$(\hat{\er}_\square(h))^{-1}\left(\left\{\tfrac{k}{m}\right\}\right)=\bigcup\limits_{\substack{I\subseteq[m]\\ |I|=k}} C_I\in \Sigma_\Z^m,$$
		where
		$C_I=\bigtimes\limits_{i\in[m]}\!C_I^{(i)}\in\Sigma_\Z^m$ with
		$$C_I^{(i)}=\left.\begin{cases}
			Z\setminus\Gamma(h)&\text{if }i\in I \\
			\Gamma(h)&\text{otherwise}
		\end{cases}\right\}\in\Sigma_\Z$$
		for $I\subseteq[m]$ and $i\in[m]$. 
		Moreover, we compute
		$$\EE_{\ul{z}\sim\DD^m}[\hat{\er}_{\ul{z}}(h)]=\frac{1}{m}\sum\limits_{i=1}^m \EE_{z\sim\DD}[\ell(h,z)]=\EE_{z\sim\DD}[\ell(h,z)]=\er_{\DD}(h). \vspace{-0.7em}$$
	\end{rem}

	Following \cite[Definition~4.3]{UnderstandingML} and \cite[\S\,3.1.2]{Vidyasagar}, we now introduce the \emph{uniform convergence property}, which guarantees that sample errors are ``good'' estimates for true errors. 
	
	\begin{defn}\label{defn::UCP}
		Suppose that there exists $m_\HH\in\NN$ such that the map 
		\begin{align*}
			U=U(\HH,m,\DD)\colon\Z^m&\to[0,1] \\
			\ul{z}&\mapsto \sup\limits_{h\in\HH}\big|\hspace{-1.25pt}\er_\DD(h)-\hat{\er}_{\ul{z}}(h)\big|
		\end{align*}
		is $\Sigma_\Z^m$--measurable for any $m\geq m_\HH$ and any $\DD\in\DDD$. Then the hypothesis space $\HH$ is said to have the \textbf{uniform convergence property (UCP)} (with respect to $\DDD$) if it satisfies the following condition: \\\vspace{-0.2cm}
		
		\begingroup
		\leftskip3.2em
		\rightskip\leftskip
		For any $\varepsilon,\delta\in(0,1)$ there exists $m_0=m_0(\varepsilon,\delta)\geq m_\HH$ such that for any $m\geq m_0$ and any $\DD\in\DDD$ the following inequality holds:
		$$\DD^m\!\left(\vphantom{\left\{\ul{z}\in\Z^m \left|\ \sup\limits_{h\in\HH}|\er_\DD(h)-\hat{\er}_{\ul{z}}(h)|\leq\varepsilon\right.\right\}}\right.\underbrace{\left\{\ul{z}\in\Z^m \left|\ \sup\limits_{h\in\HH}\big|\hspace{-1.25pt}\er_\DD(h)-\hat{\er}_{\ul{z}}(h)\big|\leq\varepsilon\right.\right\}}_{=U^{-1}([0,\varepsilon])\in\Sigma_\Z^m}\left.\vphantom{\left\{\ul{z}\in\Z^m \left|\ \sup\limits_{h\in\HH}|\er_\DD(h)-\hat{\er}_{\ul{z}}(h)|\leq\varepsilon\right.\right\}}\right)\geq 1-\delta.$$
		\endgroup
	\end{defn}
	
	The measurability of the map $U$ is not always guaranteed (see Example~\ref{exa::UCP-measurability}). The uniform convergence property can be regarded as a uniform version of the Law of Large Numbers. Roughly speaking, it ensures that, uniformly over all hypotheses in $\HH$ and over all distributions in $\DDD$, the sample error is a ``good'' approximation of the true error. To verify that $\HH$ has the uniform convergence property, we will later see that it is enough to bound the \emph{VC dimension} of $\HH$ (see Theorem~\ref{thm::VC-UCP}). The VC dimension, first introduced by and later named after Vapnik and Chervonenkis~\cite{VC1968}, measures the expressive power of a hypothesis space. The notion relies on the concept of \emph{shattering}. Given $A\subseteq\X$, we set
	$$\HH_A:=\HH\restriction_A:=\{h\restriction_A\mid h\in\HH\},$$
	and we say that $\HH$ \textbf{shatters} $A$ if $\HH_A=\{0,1\}^A$.
	
	\begin{defn}
		The \textbf{VC dimension} of $\HH$, denoted $\mathrm{vc}(\HH)$, is the maximal $d\in\NN$ such that there exists a set $A\subseteq\X$ of size $d$ that is shattered by $\HH$. If the hypothesis space $\HH$ can shatter sets of arbitrarily large size, then we say that $\HH$ has \textbf{infinite VC dimension} and write $\mathrm{vc}(\HH)=\infty$.
	\end{defn}
	
	Another related notion, also introduced in \cite{VC1968}, that can be used to measure the combinatorial capacity of a hypothesis space $\HH$ is its \textbf{growth function} $\pi_\HH\colon\NN\to\NN,$
	which is defined by
	$$\pi_\HH(m):=\max\limits_{\substack{A\subseteq\X\\ |A|=m}}|\HH_A| $$
	for $m\in\NN$. The following result allows us to polynomially bound the values of the growth function using the VC dimension. It was established independently by Sauer~\cite[Theorem~1]{Sauer}, Shelah~\cite{Shelah1972}, Vapnik and Chervonenkis~\cite[\S\,2]{VC1971}, with motivation in different areas. As Shelah~\cite[page~254]{Shelah1972} also mentions that Perles was involved in the proof, it is sometimes called the Sauer--Shelah--Perles Lemma. Others refer to the result simply as Sauer's Lemma. 
	
	\begin{lem}[Sauer, Shelah, Vapnik, Chervonenkis, Perles]\label{lem::Sauer}
		Let $d\in\NN$ and suppose that $\mathrm{vc}(\HH)\leq d$. Then 
		$$\pi_\HH(m)\leq \sum\limits_{i=1}^d \binom{m}{i}$$
		for any $m\in\NN$. In particular, $\pi_\HH(m)\leq (em/d)^d$
		for any $m>d+1$.
	\end{lem}
	
	We finally specify the formal model of learning that we work with: \emph{Probably Approximately Correct (PAC) Learning}. This concept, first introduced by Valiant~\cite{Valiant}, was modified and extended over time to also capture the agnostic setup that we deal with. The definition we use is a refined version of \cite[Definition~2.1]{Anthony-Bartlett}, \cite[Definition~3.3]{UnderstandingML} and \cite[Definition~3.5]{Vidyasagar}, further inspired by \cite[\S\,2]{Blumer}.
	
	\begin{defn}\label{defn::PAC}
		A learning function $\A$ for $\HH$ is called \textbf{probably approximately correct (PAC)} (with respect to $\DDD$) if it satisfies the following condition: \\\vspace{-0.2cm}
		
		\begingroup
		\leftskip3.2em
		\rightskip\leftskip
		For any $\varepsilon,\delta\in(0,1)$ there exists $m_0=m_0(\varepsilon,\delta)\in\NN$ such that for any $m\geq m_0$ and any $\DD\in\DDD$ there exists a set $C=C(\varepsilon,\delta,m,\DD)\in\Sigma_\Z^m$ such that 
		\begin{align*}
			&C\subseteq \left\{\ul{z}\in\Z^m\mid \er_\DD(\A(\ul{z}))-\opt_\DD(\HH)\leq\varepsilon\right\}\\
			&\text{and }\DD^m(C)\geq 1-\delta.
		\end{align*}
		\endgroup
		The hypothesis space $\HH$ is called \textbf{PAC learnable} (with respect to $\DDD$) if there exists a learning function for $\HH$ that is PAC with respect to $\DDD$.
	\end{defn}
	
	Note that the set $\{\ul{z}\in\Z^m\mid \er_\DD(\A(\ul{z}))-\opt_\DD(\HH)\leq\varepsilon\}$ is not necessarily contained in $\Sigma_\Z^m$ (see Example~\ref{exa::PAC-measurability}).\footnote{For this reason, our notion of PAC learnability employs the computation of the inner measure, which can be computed also for sets that are not measurable.} Further, we point out that in our model, PAC learnability of hypothesis spaces can be expressed with respect to any set $\DDD$ of distributions defined on the measurable space $(\Z,\Sigma_\Z)$. In the literature, however, PAC learnability is often defined in terms of the class of all (suitable) distributions (see e.g.\ \cite[Definition~2.1]{Anthony-Bartlett} \cite[Definition~3.3]{UnderstandingML}).
	
	A quite simple learning principle, referred to as \textbf{sample error minimization (SEM)}, relies on choosing hypotheses that work well on the observed data. 
	More precisely, one considers learning functions $\A$ for $\HH$ such that
	$$\hat{\er}_{\ul{z}}(\A(\ul{z}))=\hat{\opt}_{\ul{z}}(\HH)$$
	for any $m\in\NN$ and $\ul{z}\in\Z^m$. Inspired by \cite[\S\,3.3.2]{Vidyasagar}, we consider learning functions fulfilling a slightly weaker condition.
	
	\begin{defn}\label{defn::NMSE}
		A learning function $\A$ for $\HH$ is \textbf{nearly minimizing the sample error (NMSE)} if for any $\varepsilon\in(0,1)$ there exists $m_0=m_0(\varepsilon)\in\NN$ such that for any $m\geq m_0$ and any $\ul{z}\in\Z^m$ the following inequality holds:
		$$\hat{\er}_{\ul{z}}(\A(\ul{z}))-\hat{\opt}_{\ul{z}}(\HH)\leq\varepsilon.$$
	\end{defn}
	
	\section{Fundamental Theorem of Statistical Learning}\label{sec::fundthm}
	
	The aim of this section is to establish the Fundamental Theorem of Statistical Learning. This result, due to Blumer, Ehrenfeucht, Haussler and Warmuth~\cite[Theorem~2.1\:\!(i)]{Blumer}, beautifully relates the notions VC dimension and PAC learnability. More precisely, it states that, under certain broadly applicable measurability conditions, a hypothesis space is PAC learnable if and only if its VC dimension is finite. In the proofs presented in this article, we follow modern expositions such as Anthony and Bartlett~\cite{Anthony-Bartlett}, Shalev-Schwartz and Ben-David~\cite{UnderstandingML} and Vidyasagar~\cite{Vidyasagar}. While the technical steps of our arguments are largely the same as (and in parts identical to) those in these references, our emphasis is on identifying the measurability conditions required for the results to remain valid. We present a self-contained and detailed proof in a very general set-up, explicitly indicating which steps demand measurability assumptions. We thus extract the minimal measurability requirements, deliberately avoiding unnecessarily strong assumptions.
	
	Definition~\ref{defn::PAC} and Definition~\ref{defn::UCP} introduce the uniform convergence property and PAC learning with respect to a set $\DDD$ of distributions. For our version of the Fundamental Theorem of Statistical Learning, the set $\DDD$ needs to contain certain probability distributions having a fairly simple shape. 
	
	\begin{defn}
		A \textbf{discrete uniform distribution} on a measurable space $(\Omega,\Sigma)$ with $\Pot_{\fin}(\Omega)\subseteq\Sigma$ is a probability measure $\PP\colon\Sigma\to[0,1]$ of the form
		$$\PP=\sum\limits_{j=1}^\ell \frac{1}{\ell}\delta_{\omega_j},$$
		where $\ell\in\NN$ and $\omega_1,\dots,\omega_\ell\in\Omega$. 
	\end{defn}
	
	Note that for our definition of a discrete uniform distribution the underlying outcome space $\Omega$ may contain $\{\omega_1,\dots,\omega_\ell\}$ as a proper subset, in particular $\Omega$ may be (uncountably) infinite. 
	
	We proceed by specifying the measurability assumptions that we impose on the hypothesis space by describing when we call it \emph{well-behaved} (cf.\ \cite[Appendix~A1]{Blumer}). The PAC learning model considered in \cite{Blumer} relies on working with deterministic labels generated by target functions and distributions on the instance space $\X$. In contrast, we work with an agnostic model involving \emph{joint} distributions on the sample space $\Z=\X\times\{0,1\}$ (see Remark~\ref{rem::joint-distributions}, cf.\ also \cite[page~961\,f.]{Blumer}). We therefore transfer the measurability assumptions described in \cite[Appendix~A1]{Blumer} and adjust them to align with the agnostic model we consider.
	
	\begin{defn}\label{defn::well-behaved}
		Let $\X$ be a non-empty set, let $\Sigma_\Z$ be a $\sigma$--algebra on $\Z=\X\times\{0,1\}$ with $\Pot_{\fin}(\Z)\subseteq\Sigma_\Z$ and let $\DDD$ be a set of distributions on $(\Z,\Sigma_\Z)$. Then a hypothesis space $\emptyset\neq\HH\subseteq\{0,1\}^\X$ is called \textbf{well-behaved} (with respect to $\DDD$) if $\Gamma(h)\in\Sigma_\Z$ for any $h\in\HH$ and there exists $m_\HH\in\NN$ such that the map 
		\begin{align*}
			V=V(\HH,m)\colon\ \ \Z^{2m}&\to\left.\left\{\tfrac{k}{m}\ \right|\, k\in[m]_0\right\}\!, \\
			(\ul{z},\ul{z}')&\mapsto V(\ul{z},\ul{z}'):=\sup\limits_{h\in\HH}\big|\hspace{0.25pt}\hat{\er}_{\ul{z}'}(h)-\hat{\er}_{\ul{z}}(h)\big|
		\end{align*}
		is $\Sigma_\Z^{2m}$--measurable for any $m\geq m_\HH$, and the map
		\begin{align*}
			U=U(\HH,m,\DD)\colon\ \ \Z^m&\to[0,1], \\
			\ul{z}&\mapsto U(\ul{z}):=\sup\limits_{h\in\HH}\big|\hspace{-1.25pt}\er_\DD(h)-\hat{\er}_{\ul{z}}(h)\big|
		\end{align*}
		is $\Sigma_\Z^{m}$--measurable for any $m\geq m_\HH$ and any $\DD\in\DDD$.
	\end{defn}
	
	The proof of the Fundamental Theorem of Statistical Learning presented below, specifically the proof of Theorem~\ref{thm::VC-UCP-help}, relies on computing and bounding the expected values of the maps $U$ and $V$. Therefore, the measurability of these maps is necessary for our arguments to apply. In other words, the notion of well-behavedness reflects the minimal measurability requirements of the Fundamental Theorem of Statistical Learning, and thus gives rise to maximally general results. In particular, well-behavedness avoids constraints such as compactness and completeness conditions. Sufficient conditions -- such as permissibility and universal separability -- for hypothesis spaces to be well-behaved are discussed in Appendix~\ref{sec::appendix-measurability}. In fact, most hypothesis spaces arising in applications satisfy these conditions.
	
	\begin{FundThm}\label{thm::fundthm}
		Let $\X$ be a non-empty set, let $\Sigma_\Z$ be a $\sigma$--algebra on $\Z=\X\times\{0,1\}$ with $\Pot_{\fin}(\Z)\subseteq\Sigma_\Z$ and let $\DDD$ be a set of distributions on $(\Z,\Sigma_\Z)$ containing all discrete uniform distributions. Further, let $\emptyset\neq\HH\subseteq\{0,1\}^\X$ be a hypothesis space that is well-behaved with respect to $\DDD$. Then $\HH$ is PAC learnable with respect to $\DDD$ if and only if $\mathrm{vc}(\HH)<\infty$.
	\end{FundThm}
	
	At this point, we emphasize that in the literature there are several extended versions of this theorem. These entail further equivalences and quantitative aspects such as sample complexity bounds (see e.g.\ \cite[Theorem~5.5]{Anthony-Bartlett}, \cite[Theorem~2.1]{Blumer}, \cite[Theorem~6.7 and Theorem~6.8]{UnderstandingML}). The main aim for the remainder of this section is to establish the Fundamental Theorem of Statistical Learning as stated above by virtue of the following more extended version:
	
	\begin{thm}\label{thm::fundthm-conditions}
		Let $\X$ be an non-empty set, let $\Sigma_\Z$ be a $\sigma$--algebra on $\Z=\X\times\{0,1\}$ with $\Pot_{\fin}(\Z)\subseteq\Sigma_\Z$ and let $\DDD$ be a set of distributions on $(\Z,\Sigma_\Z)$ containing all discrete uniform distributions. Further, let $\emptyset\neq\HH\subseteq\{0,1\}^\X$ be a hypothesis space that is well-behaved with respect to $\DDD$. Then the following conditions are equivalent:
		\begin{enumerate}[(1)]
			\item\label{H-VC} $\HH$ has finite VC dimension.
			\item\label{H-UCP} $\HH$ has the uniform convergence property with respect to $\DDD$.
			\item\label{H-SEM} Any learning function for $\HH$ that is NMSE is PAC with respect to $\DDD$.
			\item\label{H-PAC} $\HH$ is PAC learnable with respect to $\DDD$.
		\end{enumerate}
	\end{thm}
	
	\begin{rem}\label{rem::proof-fund-thm}
		In the following we establish Theorem~\ref{thm::fundthm-conditions} by providing self-contained proofs of the implications \eqref{H-VC}$\Rightarrow$\eqref{H-UCP}, \eqref{H-UCP}$\Rightarrow$\eqref{H-SEM}, \eqref{H-SEM}$\Rightarrow$\eqref{H-PAC} and \eqref{H-PAC}$\Rightarrow$\eqref{H-VC}. Our arguments for the implications \eqref{H-SEM}$\Rightarrow$\eqref{H-PAC} and \eqref{H-PAC}$\Rightarrow$\eqref{H-VC} also apply to hypothesis spaces that are not well-behaved. However, for the uniform convergence property to be well-defined, the map $U$ from Definition~\ref{defn::UCP} has to be measurable. Furthermore, our proof of implication \eqref{H-VC}$\Rightarrow$\eqref{H-UCP} is based on the assumptions that both the maps $U$ and $V$ are random variables, as we compute and bound their expected values, respectively. For our proof of implication \eqref{H-PAC}$\Rightarrow$\eqref{H-VC}, the set $\DDD$ needs to contain all discrete uniform distributions. This assumption is not necessary for the arguments we use to establish the other implications.
	\end{rem}
	
	\subsection*{(\hyperref[H-VC]{1})$\Rightarrow$(\hyperref[H-UCP]{2})}
	
	In the following, we show that hypothesis spaces with finite VC dimension have the uniform convergence property. As a first step, we verify the uniform convergence property for hypothesis spaces $\HH$ with $\mathrm{vc}(\HH)=0$, or equivalently with $|\HH|=1$. 
	
	\begin{lem}\label{lem::Finite-H-UCP}
		Let $\X$ be a non-empty set, let $\Sigma_\Z$ be a $\sigma$--algebra on $\Z=\X\times\{0,1\}$ and let $h\in\{0,1\}^\X$ with $\Gamma(h)\in\Sigma_\Z$. Then the hypothesis space $\HH=\{h\}$ has the uniform convergence property with respect to any set $\DDD$ of distributions on $(\Z,\Sigma_\Z)$. 
	\end{lem}
	\begin{proof}
		We recall from Remark~\ref{rem::sample-error-measurable} that the map
		$\hat{\er}_\square(h)$ is $\Sigma_\Z^m$--measurable, since we assume $\Gamma(h)\in\Sigma_\Z$. In particular, all probabilities and expected values computed throughout this proof are well-defined. We can choose $m_\HH=1$, fix arbitrary $\varepsilon,\delta\in(0,1)$, set
		$$m_0=m_0(\varepsilon,\delta)=\left\lceil\frac{2\log(2/\delta)}{\varepsilon^2}\right\rceil\!\geq 1,$$
		let $m\geq m_0$ and let $\DD\in\DDD^\ast$. To verify the uniform convergence property, it suffices to justify the inequality
		$$\DD^m\left(\left\{\ul{z}\in\Z^m\left|\ \big|\hspace{-1.25pt}\er_\DD(h)-\hat{\er}_{\ul{z}}(h)\big|> \varepsilon\right.\right\}\right)\leq \delta.$$
		Furthermore, we have
		$$\er_{\DD}(h)=\EE_{z\sim\DD}[\ell(h,z)]=\EE_{\ul{z}\sim\DD^m}[\hat{\er}_{\ul{z}}(h)].$$
		We can write 
		$$\big|\hspace{-1.25pt}\er_\DD(h)-\hat{\er}_{\ul{z}}(h)\big|=\frac{1}{m}\big|Y_1(\ul{z})+\dots+Y_m(\ul{z})\big|,$$
		where 
		$$Y_i\colon\Z^m\to\left[-1,1\right],\ \ul{z}=(z_1,\dots,z_m)\mapsto \ell(h,z_i)-\er_\DD(h)$$
		is a random variable\footnote{The map $Y_i$ is a translation of the map $\theta_i\colon \Z^m\to\{0,1\},\ \ul{z}\mapsto\ell(h,z_i)$, which is $\Sigma_\Z^m$--measurable, as $\Gamma(h)\in\Sigma_\Z$ implies $$\theta_i^{-1}(\{0\})=\Z\times\dots\times\Z\times\underbrace{\Gamma(h)}_{i\text{--th position}}\times\Z\times\dots\times\Z\in\Sigma_\Z^m.$$} with $\EE_{\ul{z}\sim\DD^m}[Y_i(\ul{z})]=0$ for any $i\in[m]$. Straightforward computations show that the random variables $Y_1,\dots,Y_m$ are independent. Thus, we can apply Hoeffding's Inequality (cf.\ Pollard~\cite[Appendix~B, Corollary~3]{Pollard}) to obtain 
		\begin{align*}
			\DD^m\left(\left\{\ul{z}\in\Z^m\left|\ \big|\hspace{-1.25pt}\er_\DD(h)-\hat{\er}_{\ul{z}}(h)\big|> \varepsilon\right.\right\}\right)
			&\leq 2\exp(-m\varepsilon^2/2) \\
			&\leq 2\exp(-m_0\varepsilon^2/2) \\
			&\leq \delta.
		\end{align*}
		completing the proof.
	\end{proof}
	
	We next prove a result that shows how the growth function $\pi_\HH$ can be used to bound the probability involved in the definition of the uniform convergence property. The proof requires the hypothesis space to be well-behaved, since its core lies in bounding the expected values of the random variables $U$ and $V$. The original proof of Vapnik and Chervonenkis is also based on the assumption that these maps are measurable (see \cite[pages~265 and 268]{VC1971}). We present here a modified and adapted version of \cite[proof of Theorem~6.11]{UnderstandingML}, additionally highlighting explicitly where the required measurability assumptions come into play. While in the literature similar and even sharper bounds are established under the stronger assumption of a \emph{permissible} hypothesis space (cf.\ Haussler~\cite[Theorem~2 and Corollary~1]{Haussler1992}, Pollard~\cite[Chapter~II]{Pollard}), our result applies to the more general framework of \emph{well-behaved} hypothesis spaces, thereby extending the scope of the results beyond the permissible setting.
	
	\begin{thm}\label{thm::VC-UCP-help}
		Let $\X$ be a non-empty set, let $\Sigma_\Z$ be a $\sigma$--algebra on $\Z=\X\times\{0,1\}$, let $\DDD$ be a set of distributions on $(\Z,\Sigma_\Z)$ and let $\emptyset\neq\HH\subseteq\{0,1\}^\X$ be a hypothesis space that is well-behaved with respect to $\DDD$. Then there exists $m_\HH\in\NN$ such that for any $m\geq m_\HH$, any $\delta\in(0,1)$ and any $\DD\in\DDD$ the following inequality holds:
		$$\DD^m\left(\left\{\ul{z}\in\Z^m\left|\ \sup\limits_{h\in\HH}\big|\hspace{-1.25pt}\er_\DD(h)-\hat{\er}_{\ul{z}}(h)\big|\leq\varepsilon_0(m,\delta)\right.\right\}\right)\geq 1-\delta,$$
		where 
		$$\varepsilon_0(m,\delta)=\frac{6+2\sqrt{\log(\pi_\HH(2m))}}{\delta\sqrt{2m}}.$$
	\end{thm}
	\begin{proof}
		As $\HH$ is well-behaved, we can choose $m_\HH\in\NN$ such that the maps $U$ and $V$ from Definition~\ref{defn::well-behaved} are random variables for $m\geq m_\HH$. Let $m\geq m_\HH$, $\delta\in(0,1)$, $\DD\in\DDD$, and recall that $U=U(\HH,m,\DD)$ is given by
		\begin{align*}
			U\colon\Z^m&\to[0,1], \\
			\ul{z}&\mapsto U(\ul{z}):=\sup\limits_{h\in\HH} \big|\hspace{-1.25pt}\er_\DD(h)-\hat{\er}_{\ul{z}}(h)\big|.
		\end{align*}
		We have to show that $\DD^m(\{\ul{z}\in\Z^m\mid U(\ul{z})\leq\varepsilon_0(m,\delta)\})\geq 1-\delta$. Since the random variable $U$ is non-negative, by Markov's Inequality (see \cite[Appendix~B.1]{UnderstandingML}) it suffices to establish that
		\begin{align}
			\EE_{\ul{z}\sim\DD^m}[U(\ul{z})]\leq\delta\varepsilon_0(m,\delta). \label{EU}
		\end{align}
		To bound the expected value of $U$, we first recall from Remark~\ref{rem::sample-error-measurable} that 
		$$\er_\DD(h)=\EE_{\ul{z}'\sim\DD^m}\!\left[\hat{\er}_{\ul{z}'}(h)\right]\!.$$
		Therefore, we can write
		\begin{align*}
			\EE_{\ul{z}\sim\DD^m}[U(\ul{z})]&=\EE_{\ul{z}\sim\DD^m}\left[\sup\limits_{h\in\HH}\Big|\EE_{\ul{z}'\sim\DD^m}\!\left[\hat{\er}_{\ul{z}'}(h)\right]-\hat{\er}_{\ul{z}}(h)\Big|\right] \\
			&=\EE_{\ul{z}\sim\DD^m}\left[\sup\limits_{h\in\HH}\Big|\EE_{\ul{z}'\sim\DD^m}\!\left[\hat{\er}_{\ul{z}'}(h)-\hat{\er}_{\ul{z}}(h)\right]\!\Big|\right]\!.
		\end{align*}
		We apply Jensen's Inequality (see Bogachev~\cite[Theorem~2.12.19]{Bogachev}) to obtain
		$$\Big|\EE_{\ul{z}'\sim\DD^m}\!\left[\hat{\er}_{\ul{z}'}(h)-\hat{\er}_{\ul{z}}(h)\right]\!\Big|\leq \EE_{\ul{z}'\sim\DD^m}\Big[\big|\hat{\er}_{\ul{z}'}(h)-\hat{\er}_{\ul{z}}(h)\big|\Big].$$
		The fact that the supremum of expectation is smaller than the expectation of supremum yields
		$$\sup_{h\in\HH}\EE_{\ul{z}'\sim\DD^m}\Big[\big|\hat{\er}_{\ul{z}'}(h)-\hat{\er}_{\ul{z}}(h)\big|\Big]\leq \EE_{\ul{z}'\sim\DD^m}\!\left[V(\ul{z},\ul{z}')\right]\!,$$
		where we recall that $V=V(\HH,m)$ is given by 
		\begin{align*}
			V\colon\ \ \Z^{2m}&\to\left\{\left.\tfrac{k}{m}\ \right|\, k\in[m]_0\right\}\!, \\
			(\ul{z},\ul{z}')&\mapsto V(\ul{z},\ul{z}'):=\sup\limits_{h\in\HH}|\hspace{0.5pt}\hat{\er}_{\ul{z}'}(h)-\hat{\er}_{\ul{z}}(h)|.
		\end{align*}
		Combining these inequalities and applying Tonelli's Theorem (cf.\ Boga\-chev~\cite[Theorem~3.4.5]{Bogachev}, we obtain 
		\begin{align}
			\EE_{\ul{z}\sim\DD^m}[U(\ul{z})]&\leq \EE_{\ul{z}\sim\DD^m}\Big[\EE_{\ul{z}'\sim\DD^m}\!\left[V(\ul{z},\ul{z}')\right]\!\Big] \nonumber\\
			&=\EE_{(\ul{z},\ul{z}')\sim\DD^{2m}}\!\left[V(\ul{z},\ul{z}')\right]\!. \label{EVV}
		\end{align}
		Note that 
		$$\big|\hspace{0.5pt}\hat{\er}_{\ul{z}'}(h)-\hat{\er}_{\ul{z}}(h)\big|=\Bigg|\frac{1}{m}\sum\limits_{i=1}^m (\ell(h,z'_i)-\ell(h,z_i))\Bigg|,$$
		writing $\ul{z}=(z_1,\dots,z_m)$ and $\ul{z}'=(z'_1,\dots,z'_m)$. The expected value in \eqref{EVV} is determined based on a choice of multi-sample $(\ul{z},\ul{z}')\in\Z^{2m}$ whose components are independently distributed according to $\DD$, i.e.\ $(\ul{z},\ul{z}')$ is drawn according to $\DD^{2m}$. Therefore, the expected value in \eqref{EVV} does not change if the random sample $z_i$ and the random sample $z'_i$ are mutually exchanged. Formally, this means that the term $(\ell(h,z'_i)-\ell(h,z_i))$ is replaced by $-(\ell(h,z'_i)-\ell(h,z_i))$. Hence, for any $\ul{\sigma}=(\sigma_1,\dots,\sigma_m)\in\{\pm 1\}^m$ the expected value in \eqref{EVV} is equal to 
		$$\EE_{(\ul{z},\ul{z}')\sim\DD^{2m}}\left[\sup\limits_{h\in\HH} \Bigg|\frac{1}{m}\sum\limits_{i=1}^m \sigma_i(\ell(h,z'_i)-\ell(h,z_i))\Bigg|\right]\!.$$
		Since this holds for every $\ul{\sigma}\in\{\pm 1\}^m$, it also holds if the components of $\ul{\sigma}$ are distributed uniformly at random according to the uniform distribution over $\{\pm 1\}$, denoted $\mathcal{U}_{\pm}$. Hence, the expected value in \eqref{EVV} is also equal to
		$$\EE_{\ul{\sigma}\sim\mathcal{U}_{\pm}^m}\left[\EE_{(\ul{z},\ul{z}')\sim\DD^{2m}}\left[\sup\limits_{h\in\HH}\Bigg|\frac{1}{m}\sum\limits_{i=1}^m \sigma_i(\ell(h,z'_i)-\ell(h,z_i))\Bigg|\right]\right]\!,$$
		and by the linearity of expectation this equals
		$$\EE_{(\ul{z},\ul{z}')\sim\DD^{2m}}\left[\EE_{\ul{\sigma}\sim\mathcal{U}_{\pm}^m}\left[\sup\limits_{h\in\HH} \Bigg|\frac{1}{m}\sum\limits_{i=1}^m \sigma_i(\ell(h,z'_i)-\ell(h,z_i))\Bigg|\right]\right]\!.$$
		Now, fix $(\ul{z},\ul{z}')\in\Z^{2m}$ and let $A\subseteq\X$ be the set of the instances appearing in the multi-samples $\ul{z}$ and $\ul{z}'$, i.e.\ we set
		$$A=\{x\in\X\mid (x,y)\in\{z_i,z'_i\}\text{ for some }i\in[m]\text{ and }y\in\{0,1\}\}.$$
		We can then write
		$$\sup\limits_{h\in\HH} \Bigg|\frac{1}{m}\sum\limits_{i=1}^m \sigma_i(\ell(h,z'_i)-\ell(h,z_i))\Bigg| = \max\limits_{h\in\HH_A} \Bigg|\frac{1}{m}\sum\limits_{i=1}^m \sigma_i(\ell(h,z'_i)-\ell(h,z_i))\Bigg|.$$
		Indeed, if two hypotheses $h_1,h_2\in\HH$ coincide on the set $A$, i.e.\ $h_1\restriction_A=h_2\restriction_A$, then also $\ell(h_1,z_i)=\ell(h_2,z_i)$ and $\ell(h_1,z'_i)=\ell(h_2,z'_i)$ for any $i\in[m]$. Thus, only the finitely many functions in $\HH_A$ contribute to the computation of the supremum. Now, fixing some $h\in\HH_A$, we consider the map $V_h$ given by 
		\begin{align*}
			V_h\colon\{\pm 1\}^m&\to \RR, \\
			\ul{\sigma}&\mapsto V_h(\ul{\sigma}):=\frac{1}{m}\sum\limits_{i=1}^m \sigma_i(\ell(h,z'_i)-\ell(h,z_i)).
		\end{align*}
		The map $V_h$ is a random variable with
		$$\EE_{\ul{\sigma}\sim\mathcal{U}_{\pm}^m}\!\left[V_h(\ul{\sigma})\right]=0,$$
		and it can be written as an average of independent random variables, each of which takes values in $[-1,1]$ (see Lemma~\ref{lem::Hoeffding}). Thus, we can apply Hoeffding's Inequality (cf.\ Pollard~\cite[Appendix~B, Corollary~3]{Pollard}) to obtain 
		$$\PP(|V_h|>\rho)\leq 2\exp(-m\rho^2/2)$$
		for any $\rho>0$, writing $\PP(|V_h|>\rho)$ for $\mathcal{U}_{\pm}^m(\{\ul{\sigma}\in\{\pm 1\}^m\mid |V_h(\ul{\sigma})|>\rho\})$. Since $|A|\leq 2m$ implies $|\HH_A|\leq\pi_\HH(2m)$, this yields
		\begin{align*}
			\PP\left(\max\limits_{h\in\HH_A}|V_h|>\rho\right)&=\PP(|V_h|>\rho\text{ for some }h\in\HH_A) \\
			&\leq\sum\limits_{h\in\HH_A}\PP(|V_h|>\rho)\\
			&\leq |\HH_A|\cdot 2\exp(-m\rho^2/2) \\
			&\leq 2\pi_\HH(2m)\exp(-m\rho^2/2)
		\end{align*}
		for any $\rho>0$.
		Thus, applying Lemma~\ref{lem::PP-EE} to the random variable $\max\limits_{h\in\HH_A}|V_h|$, we obtain
		$$\EE_{\ul{\sigma}\sim\mathcal{U}_{\pm}^m}\left[\max\limits_{h\in\HH_A} \big|V_h(\ul{\sigma})\big|\right]\leq \frac{6+2\sqrt{\log(\pi_\HH(2m))}}{\sqrt{2m}}=\delta\varepsilon_0(m,\delta).$$
		Combining the above, yields the inequalities
		\begin{align}
			\begin{split}
				&\EE_{\ul{z}\sim\DD^m}[U(\ul{z})] \\
				\leq\, &\EE_{(\ul{z},\ul{z}')\sim\DD^{2m}}\!\left[V(\ul{z},\ul{z}')\right] \\
				=\;&\EE_{(\ul{z},\ul{z}')\sim\DD^{2m}}\left[\EE_{\ul{\sigma}\sim\mathcal{U}_{\pm}^m}\left[\sup\limits_{h\in\HH} \Bigg|\frac{1}{m}\sum\limits_{i=1}^m \sigma_i(\ell(h,z'_i)-\ell(h,z_i))\Bigg|\right]\right] \\
				=\;&\EE_{(\ul{z},\ul{z}')\sim\DD^{2m}}\left[\EE_{\ul{\sigma}\sim\mathcal{U}_{\pm}^m}\left[\max_{h\in\HH_A}\big|V_h(\ul{\sigma})\big|\right]\right] \\
				\leq \;&\EE_{(\ul{z},\ul{z}')\sim\DD^{2m}}\left[\delta\varepsilon_0(m,\delta)\right] \\
				=\; &\delta\varepsilon_0(m,\delta).
			\end{split}\label{EEE}
		\end{align}
		This implies the desired inequality
		$$\DD^m\left(\left\{\ul{z}\in\Z^m\left|\ \sup\limits_{h\in\HH}\big|\hspace{-1.25pt}\er_\DD(h)-\hat{\er}_{\ul{z}}(h)\big|\leq\varepsilon_0(m,\delta)\right.\right\}\right) \geq 1-\delta,$$
		and hence establishes our claim.
	\end{proof}
	
	\begin{rem}
		The proof of Theorem~\ref{thm::VC-UCP-help} relies on bounding the expected value of $U$ by the expected value of $V$ (see \eqref{EU}, \eqref{EVV} and \eqref{EEE}). Thus, assuming $\HH$ to be well-behaved is essential for Theorem~\ref{thm::VC-UCP-help}, whereas	it is not directly apparent that this assumption is necessary for establishing Theorem~\ref{thm::VC-UCP}. However, we apply Theorem~\ref{thm::VC-UCP-help} in the proof of Theorem~\ref{thm::VC-UCP}, which in turn yields the implication \eqref{H-VC}$\Rightarrow$\eqref{H-UCP}. Hence, the assumption on $\HH$ to be well-behaved in the Fundamental Theorem of Statistical Learning, arises from Theorem~\ref{thm::VC-UCP-help}.
	\end{rem}
	
	The arguments for proving the following theorem are inspired by the steps in \cite[page~51]{UnderstandingML}.
	
	\begin{thm}\label{thm::VC-UCP}
		Let $\X$ be a non-empty set, let $\Sigma_\Z$ be a $\sigma$--algebra on $\Z=\X\times\{0,1\}$ and let $\DDD$ be a set of distributions on $(\Z,\Sigma_\Z)$. Further, let
		$\emptyset\neq\HH\subseteq\{0,1\}^\X$ be a hypothesis space that is well-behaved with respect to $\DDD$. If $\HH$ has finite VC dimension, then $\HH$ has the uniform convergence property with respect to $\DDD$.
	\end{thm}
	\begin{proof}
		As $\HH$ is well-behaved, we can choose $m_\HH\in\NN$ such that the maps in Definition~\ref{defn::well-behaved} are measurable for $m\geq m_\HH$. Suppose that $d=\mathrm{vc}(\HH)<\infty$. If $d=0$, then $\HH$ is a singleton, and thus has the uniform convergence property by Lemma~\ref{lem::Finite-H-UCP}. Let $d\geq 1$, fix arbitrary $\varepsilon,\delta\in(0,1)$ and set
		$$m_0:=m_0(\varepsilon,\delta):=\left\lceil\max\left\{m_\HH, m_0^{(1)}, m_0^{(2)},m_0^{(3)}\right\}\right\rceil\!,$$
		where 
		\begin{align*}
			m_0^{(1)}&=\frac{d+1}{2}, \\
			m_0^{(2)}&=\frac{d}{2}\exp\left(\frac{9}{d}-1\right), \\
			m_0^{(3)}&=4\frac{8d}{(\delta\varepsilon)^2}\log\left(\frac{16d}{(\delta\varepsilon)^2}\right)+\Bigg|\frac{16d\log(2e/d)}{(\delta\varepsilon)^2}\Bigg|.
		\end{align*}
		We show that for any $m\geq m_0$ and any $\DD\in\DDD$ the following inequality holds:
		$$\DD^m\left(\left\{\ul{z}\in\Z^m\left|\  \sup\limits_{h\in\HH}\big|\hspace{-1.25pt}\er_\DD(h)-\hat{\er}_{\ul{z}}(h)\big|\leq\varepsilon\right.\right\}\right)\geq 1-\delta.$$
		Note that this probability is well-defined, as $m\geq m_0\geq m_\HH$ implies the measurability of the map $U(\HH,m,\DD)$ from Definition~\ref{defn::UCP}.
		Since $m\geq m_0^{(3)}$, applying \cite[Lemma~A.2]{UnderstandingML} yields
		\begin{align*}
			m&\geq \frac{8d\log(m)}{(\delta\varepsilon)^2}+\Bigg|\frac{8d\log(2e/d)}{(\delta\varepsilon)^2}\Bigg| \\
			&\geq \frac{8d\log(m)}{(\delta\varepsilon)^2}+\frac{8d\log(2e/d)}{(\delta\varepsilon)^2} =\frac{8d\log(2em/d)}{(\delta\varepsilon)^2}, 
		\end{align*}
		which is equivalent to
		\begin{align}
			\label{ineq}
			\varepsilon&\geq \frac{2\sqrt{2d\log(2em/d)}}{\delta\sqrt{m}} =\frac{4\sqrt{d\log(2em/d)}}{\delta\sqrt{2m}}.
		\end{align}
		The inequality $m\geq m_0^{(2)}$ transforms into
		$$6\leq 2\sqrt{d\log(2em/d)}.$$
		Thus, the square root in  \eqref{ineq} is real, and \eqref{ineq} implies
		\begin{align}\label{ineq2}
			\varepsilon\geq \frac{6+2\sqrt{d\log(2em/d)}}{\delta\sqrt{2m}}. 
		\end{align}
		As $2m\geq 2m_0^{(1)}=d+1$, we can apply Lemma~\ref{lem::Sauer} to obtain
		$$\pi_\HH(2m)\leq (2em/d)^d.$$
		Hence, \eqref{ineq2} implies
		\begin{align*}
			\varepsilon\geq \frac{6+2\sqrt{\log(\pi_\HH(2m))}}{\delta\sqrt{2m}}=\varepsilon_0(m,\delta),
		\end{align*}
		using the notation of Theorem~\ref{thm::VC-UCP-help}. In particular, this ensures
		\begin{align*}
			&\left\{\ul{z}\in\Z^m\left|\ \sup\limits_{h\in\HH}\big|\hspace{-1.25pt}\er_\DD(h)-\hat{\er}_{\ul{z}}(h)\big|\leq\varepsilon_0(m,\delta)\right.\right\} \\
			\subseteq\;&
			\left\{\ul{z}\in\Z^m\left|\ \sup\limits_{h\in\HH}\big|\hspace{-1.25pt}\er_\DD(h)-\hat{\er}_{\ul{z}}(h)\big|\leq\varepsilon\right.\right\}\!.
		\end{align*}
		Therefore, exploiting that $m\geq m_\HH$ and applying Theorem~\ref{thm::VC-UCP-help}, we obtain 
		\begin{align*}
			& \DD^m\left(\left\{\ul{z}\in\Z^m\left|\ \sup\limits_{h\in\HH}\big|\hspace{-1.25pt}\er_\DD(h)-\hat{\er}_{\ul{z}}(h)\big|\leq\varepsilon\right.\right\}\right) \\
			\geq\;& \DD^m\left(\left\{\ul{z}\in\Z^m\left|\ \sup\limits_{h\in\HH}\big|\hspace{-1.25pt}\er_\DD(h)-\hat{\er}_{\ul{z}}(h)\big|\leq\varepsilon_0(m,\delta)\right.\right\}\right) \\
			\geq\;&1-\delta.
			\qedhere
		\end{align*}
	\end{proof}
	
	\subsection*{(\hyperref[H-UCP]{2})$\Rightarrow$(\hyperref[H-SEM]{3})}
	
	Blumer, Ehrenfeucht, Haussler and Warmuth~\cite{Blumer} were the first to show that, given a hypothesis space $\HH$ with finite VC dimension, any learning function $\A$ for $\HH$, for which the determined hypothesis $\A(\ul{z})$ is \emph{consistent} with the underlying multi-sample $\ul{z}$, is PAC. In our framework the consistency of $\A(\ul{z})$ with $\ul{z}$ can be translated into  $\hat{\er}_{\ul{z}}(\A(\ul{z}))=0$. However, given $\ul{z}\in\Z^m$, in our agnostic setting there might not exist a hypothesis $h\in\HH$ with $\hat{\er}_{\ul{z}}(h)=0$. We recall that to this end we introduced in Definition~\ref{defn::NMSE} the notion of learning functions that are \emph{nearly minimizing the sample error} (NMSE). Adopting the computations in Vidyasagar~\cite[Theorem~3.2]{Vidyasagar}, we now show that, if a hypothesis space $\HH$ has the uniform convergence property, then any such learning function is PAC. 
	
	\begin{thm}\label{thm::UCP-NMSE}
		Let $\X$ be a non-empty set, let $\Sigma_\Z$ be a $\sigma$--algebra on $\Z=\X\times\{0,1\}$ and let $\DDD$ be a set of distributions on $(\Z,\Sigma_\Z)$. Further, let
		$\emptyset\neq\HH\subseteq\{0,1\}^\X$ be a hypothesis space that is well-behaved with respect to $\DDD$. If $\HH$ has the uniform convergence property with respect to $\DDD$, then any learning function for $\HH$ that is NMSE is PAC with respect to $\DDD$.
	\end{thm}
	\begin{proof}
		As $\HH$ is well-behaved, we can choose $m_\HH\in\NN$ such that the map $U(\HH,m,\DD)$ from Definition~\ref{defn::UCP} is measurable for any $m\geq m_\HH$ and $\DD\in\DDD$. Let $\varepsilon,\delta\in(0,1)$. Since we assume that $\HH$ has the uniform convergence property and $\A$ is NMSE, there exist $$m_0^{\mathrm{UCP}}(\tfrac{\varepsilon}{4},\delta),m_0^{\mathrm{NMSE}}(\tfrac{\varepsilon}{4})\in\NN$$ 
		such that for any 
		$$m\geq m_0:=m_0(\varepsilon,\delta):=\max\left\{m_\HH, m_0^{\mathrm{UCP}}(\tfrac{\varepsilon}{4},\delta),m_0^{\mathrm{NMSE}}(\tfrac{\varepsilon}{4})\right\},$$ 
		any  $\ul{z}\in\Z^m$ and any $\DD\in\DDD$ we have
		$$\DD^m\left(\left\{\ul{z}\in\Z^m\left|\ \sup\limits_{h\in\HH}\big|\hspace{-1.25pt}\er_\DD(h)-\hat{\er}_{\ul{z}}(h)\big|\leq\frac{\varepsilon}{4}\right.\right\}\right)\geq 1-\delta$$
		and $\hat{\er}_{\ul{z}}(\A(\ul{z}))-\hat{\opt}_{\ul{z}}(\HH)\leq\tfrac{\varepsilon}{4}$. Note that the probability is well-defined, as $m\geq m_0\geq m_\HH$ implies the measurability of $U(\HH,m,\DD)$. Fix $m\geq m_0$ and $\DD\in\DDD$. By the definition of $\opt_\DD(\HH)$, we can choose a hypothesis $h_\varepsilon\in\HH$ such that $\er_\DD(h_\varepsilon)-\opt_\DD(\HH)\leq\tfrac{\varepsilon}{4}.$
		Now let $\ul{z}\in\Z^m$ such that $\sup\limits_{h\in\HH}\big|\hspace{-1.25pt}\er_\DD(h)-\hat{\er}_{\ul{z}}(h)\big|\leq\frac{\varepsilon}{4}$ to obtain
		\begin{align}
			\forall h\in\HH\colon\; \big|\hspace{-1.25pt}\er_\DD(h)-\hat{\er}_{\ul{z}}(h)\big|\leq\frac{\varepsilon}{4}.\label{ineq5}
		\end{align}
		Together with the other inequalities from above this yields
		\begin{align*}
			\er_\DD(\A(\ul{z}))&\leq \hat{\er}_{\ul{z}}(\A(\ul{z}))+\frac{\varepsilon}{4} && (\text{\eqref{ineq5} applied to $h=\A(\ul{z})$})\\
			&\leq \hat{\opt}_{\ul{z}}(\HH)+\frac{\varepsilon}{2} && (\text{$\A$ is NMSE}) \\
			&\leq \hat{\er}_{\ul{z}}(h_\varepsilon)+\frac{\varepsilon}{2} && (\text{definition of $\hat{\opt}_{\ul{z}}(\HH)$}) \\
			&\leq \er_\DD(h_\varepsilon)+\frac{3\varepsilon}{4} && (\text{\eqref{ineq5} applied to $h=h_\varepsilon$}) \\
			&\leq \opt_\DD(\HH)+\varepsilon && (\text{choice of $h_\varepsilon$}).
		\end{align*}
		Hence, we obtain
		\begin{align*}
			C:=\; &\left\{\ul{z}\in\Z^m\left|\ \sup\limits_{h\in\HH}\big|\hspace{-1.25pt}\er_\DD(h)-\hat{\er}_{\ul{z}}(h)\big|\leq\frac{\varepsilon}{4}\right.\right\} \\
			\subseteq\; &\left\{\ul{z}\in\Z^m\mid \er_\DD(\A(\ul{z}))-\opt_\DD(\HH)\leq \varepsilon\right\}\!.
		\end{align*}
		As $C\in\Sigma_\Z^m$ and $\DD^m(C)\geq 1-\delta$, this shows that the learning function $\A$ is PAC with respect to $\DDD$.
	\end{proof}
	
	In Theorem~\ref{thm::UCP-NMSE} we assume for convenience that the hypothesis space $\HH$ is well-behaved, which ensures the measurability of the maps $U$ and $V$ from Definition~\ref{defn::well-behaved}. Note that the measurability of the map $V(\HH,m)$ is not necessary in the proof of Theorem~\ref{thm::UCP-NMSE}. 
	
	\subsection*{(\hyperref[H-SEM]{3})$\Rightarrow$(\hyperref[H-PAC]{4})}
	
	In order to prove that \eqref{H-SEM} implies \eqref{H-PAC}, it suffices to show that for any hypothesis space there exists a learning function that is NMSE.
	
	\begin{prop}\label{prop::NMSE-existence}
		Let $\X$ be a non-empty set and let $\emptyset\neq\HH\subseteq\{0,1\}^\X$ be a hypothesis space. Then there exists a learning function for $\HH$ that is SEM and thus NMSE. 
	\end{prop}
	\begin{proof}
		Set $\Z=\X\times\{0,1\}$ and let $m\in\NN$ and $\ul{z}\in\Z^m$. As we have
		$$\hat{\opt}_{\ul{z}}(\HH)=\inf\limits_{h\in\HH}\hat{\er}_{\ul{z}}(h)=\min\limits_{h\in\HH}\hat{\er}_{\ul{z}}(h),$$
		there exists $h_{\ul{z}}\in\HH$ such that $\hat{\er}_{\ul{z}}(h_{\ul{z}})=\hat{\opt}_{\ul{z}}(\HH)$. Choose such a hypothesis and set $\A(\ul{z})=h_{\ul{z}}$. Then $\A$ clearly minimizes the sample error (SEM). In particular, we obtain
		$$\hat{\er}_{\ul{z}}(\A(\ul{z}))-\hat{\opt}_{\ul{z}}(\HH)=0\leq \varepsilon$$
		for any $\varepsilon\in(0,1)$. Hence, the function $\A$ is NMSE.
	\end{proof}
	
	\subsection*{(\hyperref[H-PAC]{4})$\Rightarrow$(\hyperref[H-VC]{1})}
	
	To complete the proof of the Fundamental Theorem of Statistical Learning, we now justify that hypothesis spaces with infinite VC dimension are not PAC learnable. Our arguments are guided by the proof of the \emph{No Free Lunch Theorem} presented in \cite[Theorem~5.1]{UnderstandingML}. 
	
	\begin{thm}\label{thm::PAC-VC}
		Let $\X$ be a non-empty set, let $\Sigma_\Z$ be a $\sigma$--algebra on $\Z=\X\times\{0,1\}$ with $\Pot_{\fin}(\Z)\subseteq\Sigma_\Z$ and let $\DDD$ be a set of distributions on $(\Z,\Sigma_\Z)$ containing all discrete uniform distributions. Further, let $\emptyset\neq\HH\subseteq\{0,1\}^\X$ be a hypothesis space fulfilling $\Gamma(h)\in\Sigma_\Z$ for any $h\in\HH$. If $\HH$ is PAC learnable with respect to $\DDD$, then $\HH$ has finite VC dimension.
	\end{thm}
	\begin{proof}
		We proof the result by verifying its contrapositive. More precisely, we assume that $\HH$ has unbounded VC dimension. Based on this assumption, we justify that for any learning function
		$$\A\colon\bigcup\limits_{m\in\NN}\Z^m\to\HH$$
		and any $m\in\NN$ there exists a distribution $\DD\in\DDD$ such that any set $C'\in\Sigma_\Z^m$ with $\{\ul{z}\in\Z^m\mid \er_\DD(\A(\ul{z}))-\opt_\DD(\HH)> \tfrac{1}{8}\}\subseteq C'$ satisfies $\DD^m(C')\geq\tfrac{1}{7}$. It is straightforward to verify that this condition is sufficient for deriving that $\A$ is not PAC with respect to $\DDD$. Now, let $\A$ be an arbitrary learning function for $\HH$ and let $m\in\NN$. As $\mathrm{vc}(\HH)=\infty$, there exists a set $S\subseteq\X$ of size $2m$ that is shattered by $\HH$, i.e.\ $\{0,1\}^S=\HH_S$. We set $T:=2^{2m}=|\{0,1\}^S|$ and let $h_1,\dots, h_T\in\HH$ such that 
		$$\{0,1\}^S=\{f_1,\dots,f_T\},$$ 
		where $f_i=h_i\restriction_S$ for $i\in[T]$. For each $i\in[T]$ we define a distribution $\DD_i$ on the finite set $\Z_S=S\times\{0,1\}$, more precisely on its discrete $\sigma$--algebra $\Pot(\Z_S)$, by setting
		\begin{align}
			\DD_i(\{z\})=\begin{cases}
				\frac{1}{2m}&\text{if }z\in\Gamma(f_i), \\
				0&\text{otherwise},
			\end{cases} \label{Di}
		\end{align}
		for $z\in\Z_S$. Clearly, we have 
		\begin{align}
			\er_{\DD_i}(f_i)=\DD_i(\Z_S\setminus\Gamma(f_i))=0. \label{Di-error}
		\end{align}
		\begin{claim}[1]\label{claim1}
			The map
			\begin{align*}
				\A_S\colon \Z^m&\to\{0,1\}^S, \\
				\ul{z}&\mapsto \A(\ul{z})\restriction_S.
			\end{align*}
			fulfills
			\begin{align}
				\max\limits_{i\in[T]}\,\EE_{\ul{z}\sim\DD_i^m}[\er_{\DD_i}(\A_S(\ul{z}))]\geq\frac{1}{4}. \label{claim1-ineq}
			\end{align}
		\end{claim}
		\begin{proof}[Proof of Claim (1)]
			\renewcommand{\qedsymbol}{$\Diamond$}
			First, note that the expected values are well-defined, since the map
			$$\Z_S^m\to[0,1],\ \ul{z}\mapsto \er_{\DD_i}(\A_S(\ul{z}))$$
			is bounded and $\Pot(\Z_S^m)$--measurable for any $i\in[T]$. Next, we justify inequality~\eqref{claim1-ineq}. There are $k:=|S^m|=(2m)^m$ possible $m$--tuples of instances from $S$. We write $S^m=\{\ul{a}_1,\dots, \ul{a}_k\}$ with $\ul{a}_j\in S^m$. Further, given $j\in[k]$, $\ul{a}_j=(x_1,\dots,x_m)$ and $i\in[T]$, we write
			$$\ul{z}_j^i=((x_1,f_i(x_1)),\dots,(x_m,f_i(x_m))).$$
			For any $\ul{z}=(({x_1},y_1),\dots,({x_m},y_m))\in\Z_S^m$ and any $i\in[T]$, we compute
			\begin{align*}
				\DD_i^m(\{\ul{z}\})&=\prod\limits_{\ell=1}^m\underbrace{\DD_i(\{(x_\ell,y_\ell)\})}_{=\begin{cases}
						\tfrac{1}{2m}&\text{if }y_\ell=f_i(x_\ell), \\
						0&\text{otherwise},
				\end{cases}} \\
				&=\begin{cases}
					(\frac{1}{2m})^m=\frac{1}{k}&\text{if }\ul{z}=\ul{z}_j^i\text{ for some }j\in[k], \\
					0&\text{otherwise}.
				\end{cases}
			\end{align*}
			Therefore, for $i\in[T]$ we obtain
			\begin{align}
				\EE_{\ul{z}\sim\DD_i^m}[\er_{\DD_i}(\A_S(\ul{z}))]=\frac{1}{k}\sum\limits_{j=1}^k\er_{\DD_i}(\A_S(\ul{z}_j^i)).  \label{E-Di}
			\end{align}
			Exploiting the fact that the maximum is larger than the average and that the average is larger than the minimum, we compute
			\begin{align}
				\begin{split}
					&\max\limits_{i\in[T]}\EE_{\ul{z}\sim\DD_i^m}[\er_{\DD_i}(\A_S(\ul{z}))] \\
					=\;&\max\limits_{i\in[T]}\frac{1}{k}\sum\limits_{j=1}^k\er_{\DD_i}(\A_S(\ul{z}_j^i)) \\
					\geq\;& \frac{1}{T}\sum\limits_{i=1}^T\frac{1}{k}\sum\limits_{j=1}^k\er_{\DD_i}(\A_S(\ul{z}_j^i)) \\
					=\;& \frac{1}{k}\sum\limits_{j=1}^k\frac{1}{T}\sum\limits_{i=1}^T\er_{\DD_i}(\A_S(\ul{z}_j^i)) \\
					\geq\;& \min\limits_{j\in[k]}\frac{1}{T}\sum\limits_{i=1}^T\er_{\DD_i}(\A_S(\ul{z}_j^i)). 
				\end{split}\label{max-min}
			\end{align}
			Next, we fix some $j\in[k]$. We write ${\ul{a}_j}=({x_1},\dots,{x_m})\in S^m$ and let ${v_1},\dots,{v_p}\in S$ be pairwise distinct such that $S\setminus\{{x_1},\dots,{x_m}\}=\{{v_1},\dots,{v_p}\}$, i.e.\ $v_1,\dots,v_p$ are the instances in $S$ that do not appear in ${\ul{a}_j}$. Clearly, we have $|S|=2m\geq p\geq m$. Thus, for every function $f\in\{0,1\}^S$ and any $i\in[T]$ we compute
			\begin{align*}
				\er_{\DD_i}(f)&=\DD_i(\Z_S\setminus\Gamma(f)) \\
				&=\frac{1}{2m} \sum\limits_{{x}\in S} \underbrace{\1_{\Z_S\setminus\Gamma(f_i)}({x},f({x}))}_{=1\Leftrightarrow f({x})\neq f_i({x})} \\
				&\geq \frac{1}{2m} \sum\limits_{r=1}^p \1_{\Z_S\setminus\Gamma(f_i)}({v_r},f({v_r})) \\
				&\geq \frac{1}{2p} \sum\limits_{r=1}^p \1_{\Z_S\setminus\Gamma(f_i)}({v_r},f({v_r})).
			\end{align*}
			Hence, again exploiting the fact that the average is larger than the minimum, we obtain
			\begin{align}
				\begin{split}
					&\frac{1}{T}\sum\limits_{i=1}^T\er_{\DD_i}(\A_S(\ul{z}_j^i)) \\ 
					\geq\;& \frac{1}{T}\sum\limits_{i=1}^T \frac{1}{2p} \sum\limits_{r=1}^p \1_{\Z_S\setminus\Gamma(f_i)}({v_r},(\A_S(\ul{z}_j^i))({v_r})) \\ 
					=\;& \frac{1}{2p}\sum\limits_{r=1}^p \frac{1}{T} \sum\limits_{i=1}^T \1_{\Z_S\setminus\Gamma(f_i)}({v_r},(\A_S(\ul{z}_j^i))({v_r})) \\ 
					\geq\;& \frac{1}{2}\min\limits_{r\in[p]}\frac{1}{T}\sum\limits_{i=1}^T \1_{\Z_S\setminus\Gamma(f_i)}({v_r},(\A_S(\ul{z}_j^i))({v_r})).
				\end{split}\label{average-min}
			\end{align}
			Next, we also fix some $r\in[p]$. We can partition the functions $f_1,\dots, f_T$ into $T/2$ disjoint pairs such that for any such pair $(f_{i_t},f_{i_{t'}})$, $t\in[T/2]$, and any ${x}\in S$ we have
			$$f_{i_t}({x})\neq f_{i_{t'}}({x})\ \Leftrightarrow\ {x}={v_r}.$$
			In particular, this guarantees $\ul{z}_j^{i_t}=\ul{z}_j^{i_{t'}}$ for any $t\in[T/2]$, since ${v_r}\neq{x_\ell}$ implies $f_{i_t}({x_\ell})=f_{i_{t'}}({x_\ell})$ for any $\ell\in[m]$. Thus, setting $f:=\A_S(\ul{z}_j^{i_t})=\A_S(\ul{z}_j^{i_{t'}})$ yields
			\begin{align*}
				\1_{\Z_S\setminus\Gamma(f_{i_t})}({v_r},f({v_r}))+\1_{\Z_S\setminus\Gamma(f_{i_{t'}})}({v_r},f({v_r}))=1.
			\end{align*}
			This implies
			\begin{align}
				\sum\limits_{i=1}^T \1_{\Z_S\setminus\Gamma(f_i)}({v_r},(\A_S(\ul{z}_j^i))({v_r}))=\frac{T}{2}. \label{T2}
			\end{align}
			Combining \eqref{max-min}, \eqref{average-min} and \eqref{T2}, we obtain
			\begin{align*}
				&\max\limits_{i\in[T]}\EE_{\ul{z}\sim\DD_i^m}[\er_{\DD_i}(\A_S(\ul{z}))] \\
				\geq\;& \min\limits_{j\in[k]}\underbrace{\frac{1}{T}\sum\limits_{i=1}^T\er_{\DD_i}(\A_S(\ul{z}_j^i))}_{\geq \frac{1}{2}\underbrace{\min\limits_{r\in[p]}\frac{1}{T}\underbrace{\sum\limits_{i=1}^T \1_{\Z_S\setminus\Gamma(f_i)}({v_r},(\A_S(\ul{z}_j^i))({v_r}))}_{=\frac{T}{2}}}_{=\frac{1}{2}}} \\
				\geq\;& \frac{1}{4}.
				\qedhere
			\end{align*}
		\end{proof}
		By \eqref{claim1-ineq} we can fix $i\in[T]$ such that 
		$$\EE_{\ul{z}\sim\DD_i^m}[\er_{\DD_i}(\A_S(\ul{z}))]\geq\frac{1}{4}.$$ 
		We extend $\DD_i$ to a distribution $\hat{\DD}$ on the discrete $\sigma$--algebra $\Pot(\Z)$ by setting
		$$\hat{\DD}(C):=\DD_i(C\cap\Z_S)=\DD_i(C\cap\Gamma(f_i))$$
		for $C\subseteq\Z$. Then the restriction $\DD$ of $\hat{\DD}$ to the $\sigma$--algebra $\Sigma_\Z$ is a discrete uniform distribution, and thus a member of $\DDD$ by our assumption. Moreover, extending $\DD_i$ in this way, yields for any $h\in\HH$ that
		\begin{align}
			\begin{split}
				\er_\DD(h)=&\DD(\Z\setminus\Gamma(h))\\
				=&\DD_i((\Z\setminus\Gamma(h))\cap\Z_S)\\
				=&\DD_i(\Z_S\setminus\Gamma(h\restriction_S)) \\
				=&\er_{\DD_i}(h\restriction_S).
			\end{split} \label{D-Di-error}
		\end{align}
		Thus, $\opt_\DD(\HH)=0$, as $\er_\DD(h_i)=\er_{\DD_i}(f_i)=0$ by \eqref{Di-error}.
		\begin{claim}[2]
			We have 
			\begin{align}
				\EE_{\ul{z}\sim\hat{\DD}^m}[\er_{\DD}(\A(\ul{z}))]\geq\frac{1}{4}.
				\label{claim2-ineq}
			\end{align}
		\end{claim}
		\begin{proof}[Proof of Claim (2)]
			\renewcommand{\qedsymbol}{$\Diamond$}
			We show that
			\begin{align*}
				\EE_{\ul{z}\sim\hat{\DD}^m}[\er_\DD(\A(\ul{z}))]=\EE_{\ul{z}\sim\DD_i^m}[\er_{\DD_i}(\A_S(\ul{z}))], 
			\end{align*}
			and apply inequality~\eqref{claim1-ineq}. The expected value on the left-hand side is well-defined, since the map
			$$\Z^m\to[0,1], \ul{z}\mapsto \er_{\DD}(\A(\ul{z}))$$
			is bounded and $\Pot(\Z^m)$--measurable. Recalling our definition of $\DD_i$ in \eqref{Di}, for any $\ul{z}=(({x_1},y_1),\dots,({x_m},y_m))\in\Z^m$ we compute
			\begin{align*}
				\hat{\DD}^m(\{\ul{z}\})&=\prod\limits_{\ell=1}^m\underbrace{\hat{\DD}(\{(x_\ell,y_\ell)\})}_{=\begin{cases}
						\DD_i(\{({x_\ell},y_\ell)\})=\tfrac{1}{2m}&\text{if }{x_\ell}\in S\text{ and }y_\ell=f_i({x_\ell}), \\
						0&\text{otherwise},
				\end{cases}} \\
				&=\begin{cases}
					(\frac{1}{2m})^m=\frac{1}{k}&\text{if }\ul{z}=\ul{z}_j^i\text{ for some }j\in[k], \\
					0&\text{otherwise}.
				\end{cases}
			\end{align*}
			Note that due to \eqref{D-Di-error} we have $$\er_\DD(\A(\ul{z}))=\er_{\DD_i}(\A(\ul{z})\restriction_S)=\er_{\DD_i}(\A_S(\ul{z}))$$
			for $\ul{z}\in\Z^m$. Combining this with \eqref{E-Di}, we obtain
			\begin{align*}
				\EE_{\ul{z}\sim\hat{\DD}^m}[\er_\DD(\A(\ul{z}))]&=\frac{1}{k}\sum\limits_{j=1}^k\er_\DD(\A(\ul{z}_j^i)) \\
				&=\frac{1}{k}\sum\limits_{j=1}^k\er_{\DD_i}(\A_S(\ul{z}_j^i)) \\
				&=\EE_{\ul{z}\sim\DD_i^m}[\er_{\DD_i}(\A_S(\ul{z}))].
				\qedhere
			\end{align*}
		\end{proof}
		\begin{claim}[3]
			The distribution $\hat{\DD}$ satisfies the inequality
			$$\hat{\DD}^m(\{\ul{z}\in\Z^m\mid \er_\DD(\A(\ul{z}))-\opt_\DD(\HH)> \tfrac{1}{8}\})\geq\frac{1}{7}.$$
		\end{claim}
		\begin{proof}[Proof of Claim (3)]
			\renewcommand{\qedsymbol}{$\Diamond$}
			Recall that $\opt_\DD(\HH)=0$ and by inequality \eqref{claim2-ineq} we have
			$$\EE_{\ul{z}\sim\hat{\DD}^m}[\er_{\DD}(\A(\ul{z}))]\geq\frac{1}{4}.$$ 
			Thus, applying Markov's Inequality (\cite[Appendix~B.1]{UnderstandingML}) yields
			\begin{align*}
				&\hat{\DD}^m(\{\ul{z}\in\Z^m\mid \er_\DD(\A(\ul{z}))-\opt_\DD(\HH)> \tfrac{1}{8}\}) \\
				=\,&\hat{\DD}^m(\{\ul{z}\in\Z^m\mid \er_\DD(\A(\ul{z}))> \tfrac{1}{8}\}) \\
				\geq\,&\frac{\EE_{\ul{z}\sim\hat{\DD}^m}[\er_\DD(\A(\ul{z}))]-\tfrac{1}{8}}{1-\tfrac{1}{8}} \\
				\geq\,&\frac{\tfrac{1}{4}-\tfrac{1}{8}}{\tfrac{7}{8}} =\frac{1}{7}.
				\qedhere
			\end{align*}
		\end{proof}
		Hence, for any $C'\in\Sigma_\Z$ with $\{\ul{z}\in\Z^m\mid \er_\DD(\A(\ul{z}))-\opt_\DD(\HH)> \tfrac{1}{8}\}\subseteq C'$, the restriction $\DD$ of $\hat{\DD}$ to the $\sigma$--algebra $\Sigma_\Z$ satisfies
		\begin{align*}
			\DD^m(C')=\hat{\DD}^m(C')\geq \hat{\DD}^m(\{\ul{z}\in\Z^m\mid \er_\DD(\A(\ul{z}))-\opt_\DD(\HH)> \tfrac{1}{8}\})\geq\frac{1}{7},
		\end{align*}
		completing our proof.
	\end{proof}
	
	\begin{rem}
		We point out that for the proof of Theorem~\ref{thm::PAC-VC} not all discrete uniform distributions are required to be members of $\DDD$. In fact, the distribution $\DD\in\DDD$ we construct resembles a uniform distribution concentrated on finitely many points of the graph of a hypothesis. Thus, it is of the more specific form 
		$$\DD=\sum\limits_{j=1}^\ell \frac{1}{\ell}\delta_{(x_j,h(x_j))}$$
		for some $\ell\in\NN$, $x_1,\dots,x_\ell\in\X$ and $h\in\{0,1\}^\X$. However, for convenience, we assume that $\DDD$ contains \emph{all} discrete uniform distributions. Finally, we remark that Theorem~\ref{thm::PAC-VC} does rely on very mild measurability conditions, namely only the assumption that $\Gamma(h)\in\Sigma_\Z$ for any $h\in\HH$.
	\end{rem}
	
	\section{Connections to Model Theory}\label{sec::model-theory}
	
	In this section, we consider applications of the Fundamental Theorem of Statistical Learning to hypothesis spaces that are definable in a model-theoretic sense over first-order structures. The structures of particular interest expand linearly ordered sets, as these are naturally endowed with the order-topology. We can thus closely examine the measure-theoretic conditions that we set up in Section~\ref{sec::fundthm} for Borel $\sigma$--algebras formed by the Borel sets arising from the order-topology.
	
	First, we introduce the general model-theoretic setup and terminology. We assume some familiarity with first-order logic and refer the reader to Mar\-ker~\cite{Marker} for further details. 
	
	To specify a (first-order) language $\LL$, we simply list the collection of function, relation and constant symbols. For instance, 
	$\Lr=\{+,-,\cdot,0,1\}$ denotes the \textbf{language of rings}, where $+,-,\cdot$ are binary function symbols and $0,1$ are constant symbols, $\LL_<=\{<\}$ denotes the \textbf{language of  orderings}, where $<$ is a binary relation symbol, the \textbf{language of ordered rings} is given by $\Lor=\Lr\cup\LL_<$, and the \textbf{language of ordered exponential fields} is given by $\LL_{\exp}=\Lor\cup\{\exp\}$ with $\exp$ denoting a unary function symbol. For a language $\LL$, an $\LL$--structure $\mathcal{M}$ is given by a non-empty domain $M$ and an interpretation of each symbol in $\LL$. For instance, $\RR_{\mathrm{or}}=(\RR,+,-,\cdot,0,1,<)$ denotes the ordered field of real numbers (with standard interpretation of the symbols), and its expansion  $\RR_{\exp}=(\RR,+,-,\cdot,0,1,<,\exp)$ denotes the real exponential field with standard exponential $\exp\colon x\mapsto \mathrm{e}^x$. 
	
	When we write $\varphi(x_1,\ldots,x_n;p_1,\ldots,p_{\ell})$ (or simply $\varphi(\ul{x};\ul{p})$ using our standard tuple notation) for a (partitioned) $\LL$--formula, we indicate that the free variables of this formula are among $x_1,\ldots,x_n,p_1,\ldots,p_\ell$, and that we distinguish between the (object) variables $x_1,\ldots,x_n$ and the (parameter) variables $p_1,\ldots,p_\ell$. Given an $\LL$--structure $\mathcal{M}$, a set $A\subseteq M^n$ is called \textbf{definable} (over $\mathcal{M}$) if there are an $\LL$--formula $\varphi(x_1,\ldots,x_n;p_1,\ldots,p_\ell)$ and parameters $w_1,\ldots,w_\ell\in M$ such that $$A=\{(a_1,\ldots,a_n)\in M^n\mid \mathcal{M}\models \varphi(a_1,\ldots,a_n;w_1,\ldots,w_\ell)\}.$$
	In this case, we write $A=\varphi(\mathcal{M};\ul{w})$ and say that $\varphi(\ul{x};\ul{w})$ \textbf{defines} $A$. For convenience, we sometimes write $\ul{a}\in A$ for $\varphi(\ul{a};\ul{w})$.
	
	Within a given $\LL$--structure $\mathcal{M}$, the hypothesis spaces we consider are specified by a definable instance space $\mathcal{X}\subseteq M^n$ and an $\LL$--formula $\varphi(\ul{x};\ul{p})$. We make this precise in the following definition.
	
	\begin{defn}\label{defn::dfblhc}
		Let $n,\ell\in \NN$, let $\LL$ be a language and let $\mathcal{M}$ be an $\LL$--structure. 
		Moreover, let $\mathcal{X}\subseteq M^n$ be a non-empty set definable over $\mathcal{M}$, and let $\varphi(x_1,\ldots,x_n;p_1,\ldots,p_\ell)$ be an $\LL$--formula. Then the hypothesis space
		$\HH^\varphi\subseteq \{0,1\}^{M^n}$ is given by
		$$\HH^\varphi:=\left\{\left.\1_{\varphi(\mathcal{M};\ul{w})}\ \right|\ \ul{w}\in M^\ell \right\},$$
		and the hypothesis space $\HH ^\varphi_{\mathcal{X}}\subseteq\{0,1\}^{\mathcal{X}}$ is given by $$\HH^\varphi_{\mathcal{X}}:=\HH^\varphi\restriction_{\mathcal{X}}=\{h\restriction_{\mathcal{X}}\mid h\in \HH^\varphi\}.$$
	\end{defn}
	
	The following example illustrates that Definition~\ref{defn::dfblhc} covers hypothesis spaces computed by classical artificial neural networks. 
	
	\begin{exa}\label{exa::rexp}
		Artificial neural networks for binary classification usually compute maps from $\RR^n$ to $\{0,1\}$ that are compositions of real variables, real-valued weights and activation functions $\sigma\colon \RR\to \RR$. For instance, a two-layer real-output sigmoid network (see Anthony and Bartlett~\cite[page~5]{Anthony-Bartlett}), to which a threshold is appended, computes hypotheses of the form 
		$$h\colon \RR^n\to \{0,1\},\ \ul{x}\mapsto \1_{[0,\infty)}\brackets{\sum_{i=1}^k u_i\sigma(\ul{v}_i\cdot \ul{x}+v_{i,0})+u_0}\!,$$
		where $\ul{x}\in \RR^n$ is the input vector, $u_i\in \RR$ (for $i\in[k]_0$) are the output weights, $\ul{v}_i\in \RR^n,v_{i,0}\in\RR$ (for $i\in[k]$) are the input weights, and the activation function $\sigma\colon \RR\to\RR$ is the standard sigmoid function $$\sigma(x)=\frac{1}{1+\exp(-x)}.$$
		Fixing such an artificial neural network, for the hypothesis space $\HH$ consisting of the computed hypotheses $h\colon \RR^n\to\{0,1\}$, there exists an $\LL_{\exp}$--formula $\varphi(x_1,\ldots,x_n;p_1,\ldots,p_\ell)$ such that over $\RR_{\exp}$ we have
		$$\HH=\HH^\varphi=\left\{\left.\1_{\varphi(\RR_{\exp};\ul{w})}\ \right|\ \!\ul{w}\in \RR^\ell \right\}\!.$$
		Moreover, one can choose this $\varphi$ such that the parameter variables $p_1,\ldots,p_\ell$ represent the collection of all weights within the artificial neural network, i.e.\ $\1_{\varphi(\RR_{\exp};\ul{w})}\colon \RR^n\to\{0,1\}$ is the hypothesis that the network computes when the weights are adjusted to be $w_1,\ldots,w_\ell\in \RR$.
		
		More generally, if an artificial neural network for binary classification only uses activation functions that are parameter-free definable over $\RR_{\exp}$ (i.e.\ their graphs are definable over $\RR_{\exp}$ by a formula without parameters), then the same as above holds. Namely, for any such hypothesis class $\mathcal{H}$, there exists an $\LL_{\exp}$--formula $\varphi$ such that over $\RR_{\exp}$ we have $\HH=\HH^\varphi$. Many activation functions that are commonly used in practice are parameter-free definable over $\RR_{\exp}$. For instance, the graph of the ReLU function $\mathrm{ReLU}(x)=\max\{0,x\}$ is defined by the $\LL_{\exp}$--formula 
		$$(x<0\to y=0)\wedge (0\leq x \to y=x).\vspace{-0.7em}$$
	\end{exa}
	
	Given a language $\LL$ expanding $\LL_{<}$, and an $\LL$-structure $\mathcal{M}$ whose reduct $(M,<)$ is a linear order, recall that $\mathcal{M}$ is \textbf{o-minimal} if any set $A\subseteq M$ that is definable over $\mathcal{M}$ can be expressed as a finite union of points and open intervals (see Marker~\cite[Definition~3.1.18]{Marker}). Steinhorn~\cite[page~27]{Steinhorn} notes that
	``PAC learning is directly relevant to neural networks whose architecture defines concept classes that consist of definable families of definable sets in o-minimal structures.''
	Most crucially, due to Wilkie~\cite{Wilkie} the real exponential field $\RR_{\exp}$ is an o-minimal $\LL_{\exp}$--structure. O-minimality implies the property \emph{NIP} (see Pillay and Steinhorn~\cite[Corollary~3.10]{Pillay-Steinhorn}), which in turn ensures that hypothesis spaces of the form $\HH_{\X}^{\varphi}$ have finite VC dimension. All hypothesis spaces as in Example~\ref{exa::rexp} are therefore of finite VC dimension.
	As mentioned in the introduction, these observations were first made by Laskowski~\cite{Laskowski}. Since Laskowski's original arguments dealt with quantitative bounds for concept classes rather than hypothesis spaces over definable instance spaces $\mathcal{X}$, we present here a self-contained and more direct proof of these observations. We first introduce the relevant model-theoretic property \emph{NIP} (see Poizat~\cite[\S\,12.4]{Poizat}), which is originally due to Shelah~\cite{Shelah1971}.
	
	\begin{defn}\label{defn::nip}
		Let $\LL$ be a language, let $\mathcal{M}$ be an $\LL$--structure and let $\varphi(x_1,\ldots,x_n;p_1,\ldots,p_\ell)$ be an $\LL$--formula. Then $\varphi(\ul{x};\ul{p})$ has \textbf{NIP} (over $\M$) if there is $m\in\NN$ such that for any set $\{\ul{a}_1,\ldots,\ul{a}_m\}\subseteq M^n$ and any set $\{\ul{w}_I\mid I\subseteq [m]\}\subseteq M^\ell$, there is some $J\subseteq [m]$ such that
		$$\M\not\models \bigwedge_{i\in J} \varphi(\ul{a}_i;\ul{w}_{J}) \wedge \bigwedge_{i\in [m]\setminus J} \neg \varphi(\ul{a}_i;\ul{w}_{J}).$$
		The structure $\mathcal{M}$ has \textbf{NIP} if every $\LL$--formula has NIP over $\M$. 
	\end{defn}                                   
	
	Definition~\ref{defn::nip} expresses that a formula $\varphi(\ul{x};\ul{p})$ has NIP if there is $m\in\NN$ such that no set of size $m$ is shattered by $\varphi$ (see Simon~\cite[Remark~2.3]{Simon}).
	
	\begin{lem}\label{lem::nipvcmodeltheory}
		Let $\LL$ be a language, let $\mathcal{M}$ be an $\LL$--structure and let \linebreak $\varphi(x_1,\ldots,\allowbreak x_n;p_1,\ldots,p_\ell)$ be an $\LL$--formula. Then 
		$\varphi$ has NIP over $\M$ if and only if $\mathrm{vc}(\mathcal{H}^\varphi)<\infty$.
	\end{lem}
	
	\begin{proof}
		Let $m\in \NN$ and let $A=\{\ul{a}_1,\ldots,\ul{a}_m\}\subseteq M^n$ be a set of size $m$. It suffices to show that $\mathcal{H}^\varphi$ shatters $A$ if and only if there is a set $\{\ul{w}_I\mid I\subseteq [m]\}\linebreak\subseteq M^\ell$ such that for any $J\subseteq [m]$,
		$$\M\models \bigwedge_{i\in J} \varphi(\ul{a}_i;\ul{w}_{J}) \wedge \bigwedge_{i\in [m]\setminus J} \neg \varphi(\ul{a}_i;\ul{w}_{J}).$$
		Indeed,
		\begin{align*}
			&\mathcal{H}^{\varphi} \text{ shatters } A \\
			\Leftrightarrow\;\ & \forall J\subseteq [m]\, \exists h_J\in \HH^{\varphi}\colon h_J^{-1}(1)\cap A =\{\ul{a}_i\mid i\in J\}\\
			\Leftrightarrow\;\ & \forall J\subseteq [m]\, \exists \ul{w}_J\in M^\ell\colon \varphi(\M;\ul{w}_J) \cap A =\{\ul{a}_i\mid i\in J\}\\
			\Leftrightarrow\;\ & \forall J\subseteq [m]\, \exists \ul{w}_J\in M^\ell\colon \M\models \bigwedge_{i\in J} \varphi(\ul{a}_i;\ul{w}_{J}) \wedge\!\! \bigwedge_{i\in [m]\setminus J} \!\neg \varphi(\ul{a}_i;\ul{w}_{J}).
		\end{align*}
	\end{proof}
	
	\begin{prop}\label{prop::nipvcmodeltheory}
		Let $\LL$ be a language and let $\mathcal{M}$ be an $\LL$--structure. Then the following are equivalent:
		\begin{enumerate}[(1)]
			\item\label{prop::nipvcmodeltheory:1} $\mathcal{M}$ has NIP.
			
			\item\label{prop::nipvcmodeltheory:2}	For any $n\in \NN$, any non-empty definable set $\mathcal{X}\subseteq M^n$ and any $\LL$--formula $\varphi(x_1,\ldots,x_n;\ul{p})$, we have $\mathrm{vc}(\HH^\varphi_{\mathcal{X}})<\infty$.
			
			\item\label{prop::nipvcmodeltheory:3}	For any $\LL$--formula $\varphi(\ul{x};\ul{p})$, we have $\mathrm{vc}(\HH^\varphi)<\infty$.
		\end{enumerate}
	\end{prop}
	
	\begin{proof}
		Lemma~\ref{lem::nipvcmodeltheory} implies that \eqref{prop::nipvcmodeltheory:1} and \eqref{prop::nipvcmodeltheory:3} are equivalent. Since $M^n$ is definable over $\mathcal{M}$, \eqref{prop::nipvcmodeltheory:3} trivially follows from \eqref{prop::nipvcmodeltheory:2}. 
		Finally, \eqref{prop::nipvcmodeltheory:3} implies \eqref{prop::nipvcmodeltheory:2}, as $\mathrm{vc}(\HH_{A})\leq \mathrm{vc}(\HH)$ holds for any hypothesis space $\mathcal{H}$ and subset $A$ of its instance space.
	\end{proof}
	
	Proposition~\ref{prop::nipvcmodeltheory} closely relates the model-theoretic notion NIP with the concept of VC dimension from Statistical Learning Theory. Namely, any hypothesis space $\HH^\varphi_{\mathcal{X}}$ over an NIP structure has finite VC dimension. On the other hand, the Fundamental Theorem of Statistical Learning connects VC dimension with PAC learning. Therefore, drawing a direct connection between NIP and PAC learning suggests itself (cf.\ e.g.\ Steinhorn~\cite[page~27]{Steinhorn}). However, the measurability requirements that we carefully scrutinized in Section~\ref{sec::fundthm} have to be accounted for. 
	
	In the following, we examine when hypothesis spaces $\HH^\varphi_{\mathcal{X}}$ are PAC learnable for the special case that the underlying $\LL$--structure is an o-minimal expansion of $\RR_\mathrm{or}=(\RR,+,-,\cdot,0,1,<)$. While our measure-theoretic examination is guided by the work of Karpinski and Macintyre~\cite{Karpinski-Macintyre}, we add further adjustments to establish agnostic PAC learnability in this context. 
	
	Given $k\in\NN$, we recall that the \textbf{Borel $\sigma$--algebra} $\B(\RR^k)$ of $\RR^k$ is the smallest $\sigma$--algebra containing all open sets in $\RR^k$. The sets in $\B(\RR^k)$ are called \textbf{Borel sets}. Note that $\B(\RR^k)=\bigotimes_{i=1}^k \B(\RR)$ (cf.\ Bogachev~\cite[Lemma~6.4.2]{Bogachev2}). For any $\mathcal{Y}\subseteq\RR^k$, we consider the \textbf{trace $\sigma$--algebra} given by $$\B(\mathcal{Y}):=\B(\RR^k)\cap{\mathcal{Y}}:=\{B\cap \mathcal{Y}\mid B\in\B(\RR^k)\}.$$ Note that $\Pot_{\fin}(\mathcal{Y})\subseteq\B(\mathcal{Y})$, and $\B(\mathcal{Y})=\{B\in\B(\RR^k)\mid B\subseteq\mathcal{Y}\}\subseteq\B(\RR^k)$ if and only if $\mathcal{Y}\in\B(\RR^k)$.
	
	\begin{lem}\label{lem::Borel}
		Let $\LL$ be a language expanding $\Lor$, let $\mathcal{R}$ be an o-minimal $\LL$--expansion of $\RR_{\mathrm{or}}$, let $n, \ell\in \NN$, let $\varphi(x_1,\ldots,x_n;p_1,\dots,p_\ell)$ be an $\LL$--formula and let $\ul{w}\in\RR^\ell$. Then $\varphi(\mathcal{R};\ul{w})\in \B(\RR^n)$.\footnote{In Kaiser~\cite[Proposition~1.1]{Kaiser} a similar result is stated and proved.}
	\end{lem}
	\begin{proof}
		Due to the Cell Decomposition Theorem (van den Dries~\cite[page~52]{Van-den-Dries}), the set $\varphi(\mathcal{R};\ul{w})$ has a finite partition into cells (see Definition~\ref{defn::cell}). It is readily verified that cells are Borel sets (see Lemma~\ref{lem::cells-Borel}), and hence $\varphi(\mathcal{R};\ul{w})\in\B(\RR^n)$.
	\end{proof}
	
	\begin{thm}\label{thm::o-minimal-PAC}
		Let $\LL$ be a language expanding $\Lor$, let $\mathcal{R}$ be an o-minimal $\LL$--expansion of $\RR_{\mathrm{or}}$, let $n, \ell\in \NN$, let $\mathcal{X}\subseteq \RR^n$ be a non-empty set that is definable over $\mathcal{R}$, and let $\varphi(x_1,\ldots,x_n;p_1,\dots,p_\ell)$ be an $\LL$--formula. Further, let $\Sigma_\Z$ be a $\sigma$--algebra on $\Z=\X\times\{0,1\}$ with $\B(\Z)\subseteq\Sigma_\Z$, and let $\DDD$ be a set of distributions on $(\Z,\Sigma_\Z)$ such that $(\Z^m,\Sigma_\Z^m,\DD^m)$ is a complete\footnote{A probability space $(\Omega,\Sigma,\PP)$ is called \textbf{complete} if $\Sigma$ contains all subsets of every set $N\in\Sigma$ with $\PP(N)=0$.} probability space for any $\DD\in\DDD$ and any $m\in\NN$. Then $\HH_\X^\varphi$ is PAC learnable with respect to $\DDD$.
	\end{thm}
	\begin{proof}
		As an o-minimal $\LL$--structure, $\mathcal{R}$ has NIP. Thus, it follows from Proposition~\ref{prop::nipvcmodeltheory} that $\mathrm{vc}(\HH_\X^\varphi)<\infty$. We aim at applying the Fundamental Theorem of Statistical Learning to deduce that the hypothesis space $\HH_\X^\varphi$ is PAC learnable.\footnote{For this conclusion it is not necessary to guarantee that $\DDD$ contains all discrete uniform distributions (see Remark~\ref{rem::proof-fund-thm}).} To this end, we have to verify that $\HH_\X^\varphi$ is well-behaved with respect to $\DDD$. 
		First note that $\Z=\X\times\{0,1\}$ is definable over $\R$, since $\X$ is definable over $\R$. For any $\ul{w}\in\RR^\ell$ we set $h_{\ul{w}}:=\1_{\varphi(\R;\ul{w})}\restriction_\X$. The graph $\Gamma(h_{\ul{w}})\subseteq\Z$ is defined over $\R$ by
		$\ul{x}\in\X\land \gamma(z;\ul{w})$, where $z=(\ul{x},y)$ and $\gamma(z;\ul{w})$ is the $\LL$--formula
		$$(\varphi(\ul{x};\ul{w})\rightarrow y=1)\land (\neg\:\!\varphi(\ul{x};\ul{w})\rightarrow y=0),$$
		and hence $\Gamma(h_{\ul{w}})\in\B(\Z)\subseteq\Sigma_\Z$ by Lemma~\ref{lem::Borel}. \pagebreak
		
		In the following, we verify that the maps $U(\HH_\X^\varphi,m,\DD)$ and $V(\HH_\X^\varphi,m)$ from Definition~\ref{defn::well-behaved} are $\Sigma_\Z^m$--measurable, respectively $\Sigma_\Z^{2m}$--measurable, for any $m\in\NN$ and any $\DD\in\DDD$. To this end, we first show that the map 
		$$f_U\colon\Z^m\times\RR^\ell\to [0,1],\ (\ul{z};\ul{w})\mapsto \big|\hspace{-1.25pt}\er_\DD(h_{\ul{w}})-\hat{\er}_{\ul{z}}(h_{\ul{w}})\big|$$
		is $(\Sigma_\Z^m\otimes\B(\RR^{\ell}))$--measurable, that the map
		$$f_V\colon\Z^{2m}\times\RR^\ell\to\left\{\left.\tfrac{k}{m}\ \right|\, k\in[m]_0\right\},\ (\ul{z},\ul{z}';\ul{w})\mapsto \big|\hat{\er}_{\ul{z}'}(h_{\ul{w}})-\hat{\er}_{\ul{z}}(h_{\ul{w}})\big|$$
		is $(\Sigma_\Z^{2m}\otimes\B(\RR^\ell))$--measurable, and then we exploit the completeness of $(\Z^m,\Sigma_\Z^m,\DD^m)$ respectively $(\Z^{2m},\Sigma_\Z^{2m},\DD^{2m})$. \\
		As $\Z$ and $\Gamma(h_{\ul{w}})$ are definable over $\R$ for any $\ul{w}\in\RR^\ell$, also the set
		$$M=\{(z;\ul{w})\in \RR^{n+1+\ell}\mid z\in\Z\setminus\Gamma(h_{\ul{w}}) \}\subseteq \Z\times\RR^\ell$$
		is definable over $\R$, and thus $M\in\B(\Z\times\RR^\ell)$ by Lemma~\ref{lem::Borel}. As $\B(\RR^{n+1+\ell})=\B(\RR^{n+1})\otimes\B(\RR^\ell)$, we can apply Lemma~\ref{lem::product-trace} to obtain $M\in\B(\Z\times\RR^\ell)=\B(\Z)\otimes\B(\RR^\ell)$. It then follows from Bogachev~\cite[Proposition~3.3.2\,(ii)]{Bogachev} that the map
		\begin{align*}
			\er_\DD(h_\square)\colon \RR^\ell&\to [0,1], \\
			\ul{w}&\mapsto \er_\DD(h_{\ul{w}})=\DD(\Z\setminus\Gamma(h_{\ul{w}}))=\DD(\{z\in\Z\mid (z;\ul{w})\in M\})
		\end{align*}
		is $\B(\RR^\ell)$--measurable. Next, we consider the map
		\begin{align*}
			\hat{\er}_\square(h_\square)\colon \Z^m\times\RR^\ell&\to \left\{\left.\tfrac{k}{m}\ \right|\, k\in[m]_0\right\}, \\
			(\ul{z};\ul{w})&\mapsto \hat{\er}_{\ul{z}}(h_{\ul{w}}).
		\end{align*}
		For any $k\in[m]_0$ the preimage of the interval $\left[\tfrac{k}{m},\infty\right)\subseteq\RR$ under this map
		$$P_k=\left\{\left.\!(\ul{z};\ul{w})\in\Z^m\times\RR^\ell\ \right|\ \hat{\er}_{\ul{z}}(h_{\ul{w}})\geq\tfrac{k}{m}\right\}$$
		is defined over $\R$ by the $\LL$--formula $\psi_k(z_1,\dots,z_m;\ul{w})$ given by
		$$\bigwedge\limits_{j\in[m]} z_j\in\Z \land \bigvee\limits_{\substack{I\subseteq[m]\\ |I|=k}}\bigwedge\limits_{i\in I}(z_i;\ul{w})\in M.$$
		We apply Lemma~\ref{lem::Borel} and Lemma~\ref{lem::product-trace} to obtain
		$$P_k\in\B(\Z^m\times\RR^\ell)=\B(\Z)^m\otimes\B(\RR^\ell)\subseteq\Sigma_\Z^m\otimes\B(\RR^\ell).$$ 
		Thus, the map $\hat{\er}_\square(h_\square)$ is $(\Sigma_\Z^m\otimes\B(\RR^\ell))$--measurable. The measurability of the maps $\er_\DD(h_\square)$ and $\hat{\er}_\square(_\square)$ implies that the maps $f_U$ and $f_V$ are measurable. Since we assume that the probability space $(\Z^k,\Sigma_\Z^k,\DD^k)$ is complete for any $k\in\NN$, we can apply the arguments\footnote{Pollard~\cite{Pollard} developed these arguments in the context of \emph{permissible} function classes (see Appendix~\ref{sec::appendix-measurability}).} of Pollard~\cite[page~197]{Pollard} to deduce that the map
		$$U\colon\Z^m\to[0,1],\ \ul{z}\mapsto \sup\limits_{\ul{w}\in\RR^\ell}f_U(\ul{z};\ul{w})$$
		is $\Sigma_\Z^m$--measurable and that the map
		$$V\colon\Z^{2m}\to \left\{\left.\tfrac{k}{m}\ \right|\, k\in[m]_0\right\},\ (\ul{z},\ul{z}')\mapsto \sup\limits_{\ul{w}\in\RR^\ell} f_V(\ul{z},\ul{z}';\ul{w})$$
		is $\Sigma_\Z^{2m}$--measurable. Hence, $\HH_\X^\varphi$ is well-behaved with respect to $\DDD$, and applying the Fundamental Theorem of Statistical Learning yields that $\HH_\X^\varphi$ is PAC learnable with respect to $\DDD$.
	\end{proof}
	
	In the final remark of this section we discuss the extent and applicability of Theorem~\ref{thm::o-minimal-PAC}.
	
	\begin{rem}\
		\begin{enumerate}[(a)]
			\item For deducing the well-behavedness of hypothesis spaces of the form $\HH_\X^\varphi$, the most crucial condition in Theorem~\ref{thm::o-minimal-PAC} is that $(\Z^m,\Sigma_\Z^m,\DD^m)$ is a complete probability space for any $m\in\NN$ and any $\DD\in\DDD$. This is trivially satisfied if $\Z$ is countable, as then $\Pot(\Z)=\B(\Z)=\Sigma_\Z$. Another sufficient condition for well-behavedness of hypothesis spaces is universal separability (see Appendix~\ref{sec::appendix-measurability} for further details).
			\item Given an arbitrary probability space $(\Omega,\Sigma,\PP)$, there always exists a (Lebesgue) completion, i.e.\ a probability space $(\Omega,\hat{\Sigma},\hat{\PP})$ that is complete and fulfills $\Sigma\subseteq\hat{\Sigma}$ as well as $\hat{\PP}\restriction_\Sigma=\PP$ (cf.\ Bogachev~\cite[page~22]{Bogachev}). However, it is well-known that the product of two complete probability spaces is in general not complete. Therefore, considering the completion $(\Z,\hat{\Sigma}_\Z,\hat{\DD})$ does not guarantee that also the product space $(\Z^m,(\hat{\Sigma}_\Z)^m,(\hat{\DD})^m)$ is complete for any $m\in\NN$.  \\
			A possible solution to this issue would be to modify Definition~\ref{defn::well-behaved} by replacing the probability space $(\Z^m,\Sigma_\Z^m,\DD^m)$ with its respective completion at the relevant places for each $m\in\NN$ and each $\DD\in\DDD$. Additionally, other definitions, such as Definition~\ref{defn::UCP} and Definition~\ref{defn::PAC}, would need to be adjusted accordingly. Hence, this approach could potentially result in a new version of Theorem~\ref{thm::fundthm-conditions}, which would primarily require the hypothesis space to be well-behaved with respect to the completions. 
			\item In the context of Theorem~\ref{thm::o-minimal-PAC}, the measurability of the map $V=V(\HH_\X^\varphi,m)$ can be established without the completeness condition. In fact, for any $k\in[m]_0$ and any $(\ul{z},\ul{z}')\in\Z^{2m}$, we can write
			\begin{align*}
				V(\ul{z},\ul{z}')\geq\tfrac{k}{m}\ \Leftrightarrow\ \big|\hat{\er}_{\ul{z}'}(h_{\ul{w}})-\hat{\er}_{\ul{z}}(h_{\ul{w}})\big|\geq\tfrac{k}{m}\text{ for some }\ul{w}\in\RR^\ell.
			\end{align*}
			Moreover, we note that the map $\hat{\er}_\square(h_\square)$ is $\LL$--definable over $\R$. Based on this, one can easily show that also the preimage of each interval $\big[\tfrac{k}{m},\infty\big)$ under $V$ is $\LL$--definable over $\R$. Hence, one can apply Lemma~\ref{lem::Borel} to deduce the measurability of $V$. \\
			Applying similar arguments to demonstrate the measurability of the map $U=U(\HH_\X^\varphi,m,\DD)$ would require dealing with sets of the form
			$$\{\ul{w}\in\RR^\ell\mid \er_\DD(h_{\ul{w}})\geq\alpha\}=\{\ul{w}\in\RR^\ell\mid \DD(\Z\setminus\Gamma(h_{\ul{w}}))\geq\alpha\}$$
			with $\alpha\in(0,1)$, which are a priori not $\LL$--definable. 
			\item In Example~\ref{exa::UCP-measurability-2} we consider a hypothesis space $\HH^\varphi$ specified by a very simple formula $\varphi$ over the reals. Our goal is to verify well-behavedness of $\HH^\varphi$ with respect to the set $\DDD^\ast$ of all distributions defined on a certain Borel $\sigma$--algebra. The set $\DDD^\ast$ includes distributions that do not satisfy the completeness condition in Theorem~\ref{thm::o-minimal-PAC} (see Example~\ref{exa::complete}). Consequently,  Theorem~\ref{thm::o-minimal-PAC} is not applicable, and the well-behavedness of $\HH$ is not a priori evident. In fact, proving the measurability of the maps $U$ and $V$ directly requires very technical and complex arguments. Thus, this example underscores the need for further results that provide sufficient conditions for the well-behavedness of hypothesis spaces. 
			
			\item Theorem~\ref{thm::o-minimal-PAC} can potentially be extended to other structures that are, in some sense, tame (e.g.\ o-minimal structures, structures with quantifier elimination) and naturally endowed with a topology (cf.\ \cite[\S\,5]{Karpinski-Macintyre}). However, one must carefully examine whether the arguments of Pollard~\cite[page~197]{Pollard} -- or those upon which Pollard's approach relies (cf.\ Dellacherie and Meyer~\cite[Chapter~III]{DM}) -- can still be applied in such settings. These arguments, for instance, require the topological space to be metrizable, separable and locally compact.
		\end{enumerate}
		\vspace{-0.7em}
	\end{rem}
	
	\section{Further Work}\label{sec::further-work}
	
	In the Fundamental Theorem of Statistical Learning the requirement that the hypothesis space $\HH$ is well-behaved with respect to the considered distribution set $\DDD$ is crucial in order to deduce that finiteness of its VC dimension implies its PAC learnability with respect to $\DDD$. However, Example~\ref{exa::UCP-measurability} presents a hypothesis space that is not well-behaved with respect to a singleton $\{\DD\}$, but has finite VC dimension and is PAC learnable with respect to $\{\DD\}$. This gives rise to the following question:
	
	\begin{frag}\label{frag::failure}
		Are there a non-empty set $\X$, a $\sigma$--algebra $\Sigma_\Z$ on $\Z=$\linebreak 
		$\X\times\{0,1\}$ with $\Pot_{\fin}(\Z)\subseteq\Sigma_\Z$, a set $\DDD$ of distributions on $(\Z,\Sigma_\Z)$ and a hypothesis space $\emptyset\neq\HH\subseteq\{0,1\}^\X$ fulfilling $\Gamma(h)\in\Sigma_\Z$ for any $h\in\HH$ such that $\HH$ has finite VC dimension but is not PAC learnable with respect to~$\DDD$?
	\end{frag}
	
	If such a pair $(\HH,\DDD)$ as in Question~\ref{frag::failure} exists, then $\HH$ is certainly not well-behaved with respect to $\DDD$. Blumer, Ehrenfeucht, Haussler and Warmuth~\cite[Appendix~A1]{Blumer} present an example of a hypothesis space that has finite VC dimension and is not PAC learnable in terms of the learning model they consider. Adapting their example to the agnostic setting could be a starting point towards answering Question~\ref{frag::failure} affirmatively. 
	
	Specializing Question~\ref{frag::failure} to the case where $\X=\RR$ and the hypothesis space is of the form $\HH^\varphi$ for some $\Lor$--formula $\varphi$, yields the following question:
	
	\begin{frag}\label{frag::failure-easy}
		Are there an $\Lor$--formula $\varphi(x;p)$ and a set $\DDD$ of distributions on $(\Z,\B(\Z))$, where $\Z=\RR\times\{0,1\}$, such that the hypothesis space
		$$\HH^\varphi=\left\{\left.\1_{\varphi(\mathcal{\RR_{\mathrm{or}}};w)}\ \right|\ \!w\in \RR \right\}$$
		is not PAC learnable?
	\end{frag}
	
	If such a pair $(\varphi,\DDD)$ as in Question~\ref{frag::failure-easy} exists, then $\HH^\varphi$ is certainly not well-behaved with respect to $\DDD$. We justify in Section~\ref{sec::model-theory} that hypothesis spaces of this form have finite VC dimension. Question~\ref{frag::failure-easy} can also be extended in several regards, namely one can consider any o-minimal expansion of $\RR_{\mathrm{or}}$, any multi-variate formula of the form $\varphi(x_1,\dots,x_n;p_1,\dots,p_{\ell})$, any definable instance space $\X\subseteq \RR^n$, and any $\sigma$--algebra on $\Z=\X\times\{0,1\}$ containing $\B(\Z)$. Further, one can ask whether $\HH^{\varphi}_{\X}$ is well-behaved with respect to $\DDD$, rather than just PAC learnable.

	
	\begingroup
	\renewcommand{\bibname}{References}

	\endgroup
	
	
	\appendix
	
	\section[Appendix: Lemmas, Examples and Further Considerations (by L.~Wirth)]{Appendix: Lemmas, Examples and Further\linebreak Considerations (by L.~Wirth)}
	
	\subsection{Auxiliary Lemmas}
	
	\begin{lem}\label{lem::Hoeffding}
		Let $\X$ be a non-empty set and consider $\Z=\X\times\{0,1\}$. Further, let $(\ul{z},\ul{z}')\in\Z^{2m}$, let $A$ be the set of instances appearing in the multi-samples $\ul{z}$ and $\ul{z}'$ and let $h\in\HH_A$. Then the map $V_h$ given by
		\begin{align*}
			V_h\colon \{\pm 1\}^m&\to\RR, \\
			\ul{\sigma}=(\sigma_1,\dots,\sigma_m)&\mapsto \frac{1}{m}\sum\limits_{i=1}^m\sigma_i(\ell(h,z'_i)-\ell(h,z_i))
		\end{align*}
		is a random variable on the discrete measurable space $(\{\pm 1\}^m,\Pot(\{\pm 1\}^m))$ with 
		$$\EE_{\ul{\sigma}\sim\mathcal{U}_{\pm}^m}[V_h(\ul{\sigma})]=0.$$
		Furthermore, $V_h$ is an average of independent random variables, each of which takes values in $[-1,1]$.
	\end{lem}
	\begin{proof}
		We can write
		$$V_h=\frac{1}{m}\sum\limits_{i=1}^m V_h^{(i)},$$
		where, for any $i\in[m]$, the random variable\footnote{Note that the map $V_h^{(i)}$ is measurable, as it is defined on the discrete measurable space $(\{\pm 1\}^m,\Pot(\{\pm 1\}^m))$.} $V_h^{(i)}$ is given by 
		\begin{align*}
			V_h^{(i)}\colon\{\pm 1\}^m&\to\{-1,0,1\}\subseteq\RR, \\
			\ul{\sigma}&\mapsto V_h^{(i)}(\ul{\sigma}):=\sigma_i\Delta_h^{(i)}
		\end{align*}
		with $\Delta_h^{(i)}=\ell(h,z'_i)-\ell(h,z_i)\in\{-1,0,1\}$. We have
		\begin{align*}
			\PP(V_h^{(i)}=0)&=\begin{cases}
				1&\text{if }\Delta_h^{(i)}=0, \\
				0&\text{otherwise},
			\end{cases} \\
			\PP(V_h^{(i)}=\ell)&=\begin{cases}
				\frac{1}{2}&\text{if }\Delta_h^{(i)}\neq0, \\
				0&\text{otherwise},
			\end{cases} \quad\quad(\ell\in\{\pm1\}),
		\end{align*}
		writing $\PP(V_h^{(i)}=k)$ for $\U_{\pm}^m(\{\ul{\sigma}\in\{\pm1\}^m\mid V_h^{(i)}(\ul{\sigma})=k\})$, $k\in\{-1,0,1\}$. Straightforward computations show that $V_h^{(1)},\dots, V_h^{(m)}$ are independent. Moreover, we have
		$$\EE_{\ul{\sigma}\sim\mathcal{U}_{\pm}^m}[V_h(\ul{\sigma})]=\frac{1}{m}\sum\limits_{i=1}^m \EE_{\ul{\sigma}\sim\mathcal{U}_{\pm}^m}[V_h^{(i)}(\ul{\sigma})]=0, $$
		since
		\begin{align*}
			\EE_{\ul{\sigma}\sim\mathcal{U}_{\pm}^m}[V_h^{(i)}(\ul{\sigma})]&=\EE_{\sigma_i\sim\U_{\pm}}[\sigma_i\Delta_h^{(i)}]=\frac{1}{2}\Delta_h^{(i)} -\frac{1}{2}\Delta_h^{(i)}=0
		\end{align*}
		for any $i\in[m]$.
	\end{proof}
	
	The following result is a refined version of Shalev-Shwartz and Ben-David~\cite[Lemma~A.4]{UnderstandingML}.
	
	\begin{lem}\label{lem::PP-EE}
		Let $(\Omega,\Sigma,\PP)$ be a probability space, let $X$ be a non-negative random variable and assume that there exist $\alpha>0$ and $\beta>0$ such that for any $\rho> 0$ we have 
		$$\PP(X>\rho)\leq 2\beta\exp(-\rho^2/\alpha^2).$$ 
		Then 
		$$\EE[X]\leq \alpha(3+\sqrt{\log(\beta)}).$$
	\end{lem}
	\begin{proof}
		If $\beta\geq e$, then applying \cite[Lemma~A.4]{UnderstandingML} (with $x'=0, a=\alpha, b=\beta$) yields
		$$\EE[X]\leq \alpha(2+\sqrt{\log(\beta)}).$$
		Otherwise, applying \cite[Lemma~A.4]{UnderstandingML} (with $x'=0, a=\alpha, b= e$) yields 
		$$\EE[X]\leq \alpha(2+\sqrt{\log(e)})=3\alpha.$$
		Both inequalities imply our claim.
	\end{proof}
	
	Given a measurable space $(\Omega,\Sigma)$ and a set $A\subseteq\Omega$, we denote by $\Sigma\cap A:=\{S\cap A\mid S\in\Sigma\}$ the trace $\sigma$--algebra on $A$.
	
	\begin{lem}\label{lem::product-trace}
		Let $(\Omega_1,\Sigma_1)$,$(\Omega_2,\Sigma_2)$ be measurable spaces and let $A_1\subseteq\Omega_1$, $A_2\subseteq\Omega_2$. Then we have $(\Sigma_1\otimes\Sigma_2)\cap(A_1\times A_2)=(\Sigma_1\cap A_1)\otimes(\Sigma_2\cap A_2)$.
	\end{lem}
	\begin{proof}
		The product $\sigma$--algebra $\Sigma_1\otimes\Sigma_2$ is generated by the system $\{E_1\times E_2\mid E_1\in\Sigma_1, E_2\in\Sigma_2\}$. Thus, it follows from Bogachev~\cite[Corollary~1.2.9]{Bogachev} that the trace $\sigma$--algebra $(\Sigma_1\otimes\Sigma_2)\cap(A_1\times A_2)$ is generated by the system 
		\begin{align*}
			&\{(E_1\times E_2)\cap(A_1\times A_2)\mid E_1\in\Sigma_1, E_2\in\Sigma_2\} \\
			=\;&\{(E_1\cap A_1)\times (E_2\cap A_2)\mid E_1\in\Sigma_1, E_2\in\Sigma_2\} \\
			=\;&\{E'_1\times E'_2\mid E'_1\in (\Sigma_1\cap A_1), E'_2\in(\Sigma_2\cap A_2)\},
		\end{align*}
		which also generates the product $\sigma$--algebra $(\Sigma_1\cap A_1)\otimes(\Sigma_2\cap A_2)$. Hence, the $\sigma$--algebras coincide.
	\end{proof}
	
	\subsection{Well-Behaved Hypothesis Spaces}\label{sec::appendix-measurability}
	
	As the notion of a well-behaved hypothesis space is quite elusive, in this section we discuss cases in which well-behavedness is partly or fully satisfied. In particular, we present several sufficient conditions for the measurability of the maps $U(\HH,m,\DD)$ and $V(\HH,m)$ from Definition~\ref{defn::well-behaved}.
	
	\begin{rem}\label{rem::sup-countability}
		A very simple sufficient condition for the measurability of the maps $U(\HH,m,\DD)$ and $V(\HH,m)$ is countability of the involved sets. Indeed, if $\X$ and $\Z$ are countable, then $\Sigma_\Z$ coincides with the discrete $\sigma$--algebra $\Pot(\Z)$, as we usually assume that $\Pot_{\fin}(\Z)\subseteq\Sigma_\Z$. Hence, every set and every function are measurable. If the hypothesis space $\HH$ is countable, then both the maps $U$ and $V$ take suprema over a countable family of measurable functions. In fact, our standard assumption that $\Gamma(h)\in\Sigma_\Z$ for any $h\in\HH$ and our considerations in Remark~\ref{rem::sample-error-measurable} imply that the map
		\begin{align*}
			\Z^m&\to[0,1], \\
			\ul{z}&\mapsto\big|\hspace{-1.25pt}\er_\DD(h)-\hat{\er}_{\ul{z}}(h)\big|
		\end{align*}
		is $\Sigma_\Z^m$--measurable and that the map
		\begin{align*}
			\Z^{2m}&\to\left\{\left.\tfrac{k}{m}\ \right| k\in[m]_0\right\}, \\
			(\ul{z},\ul{z}')&\mapsto \big|\hspace{0.25pt}\hat{\er}_{\ul{z}'}(h)-\hat{\er}_{\ul{z}}(h)\big|
		\end{align*}
		is $\Sigma_\Z^{2m}$--measurable. In general, we however need to take suprema over uncountable families of these measurable functions. In this situation, verifying the measurability of the maps $U$ and $V$ is more complicated or even impossible (see Example~\ref{exa::UCP-measurability}). However, in this section we aim at identifying further sufficient conditions that might be helpful in the uncountable setting as well.
	\end{rem}
	
	We now introduce the concept of a \emph{universally separable} hypothesis space. Dudley~\cite[page~902]{Dudley1978} was the first to introduce this property ``as a way of avoiding measurability difficulties'' (Pollard~\cite[page~38]{Pollard}).
	
	\begin{defn}\label{defn::universally-separable}
		Let $\X$ be an arbitrary set. A hypothesis space $\HH\subseteq\{0,1\}^\X$ is called \textbf{universally separable} if there exists a countable subset $\HH_0\subseteq\HH$ fulfilling the following condition: \\\vspace{-0.2cm}
		
		\begingroup
		\leftskip3.2em
		\rightskip\leftskip
		
		For any $h\in\HH$ there exists a sequence $\{h_n\}_{n\in\NN}\subseteq\HH_0$ such that for any $x\in\X$ there is $n_x\in\NN$ such that $h(x)=h_n(x)$ for any $n\geq n_x$, i.e.\ $\{h_n\}_{n\in\NN}$ converges pointwise to $h$.
		
		\endgroup
	\end{defn}
	
	We next prove that universal separability ensures measurability of the maps $U$ and $V$ from Definition~\ref{defn::well-behaved}. Similar statements in a more general setting can be found in the literature (cf.\ e.g.\ Ben-David, Mansour and Benedek~\cite[Lemma~5.2]{Ben-David-Benedek-Mansour} and Pollard~\cite[Chapter~II, Problem~3]{Pollard}). Guided by the proof of \cite[Lemma~5.2]{Ben-David-Benedek-Mansour} and by arguments suggested by D.~Pollard\footnote{The second author thanks D.~Pollard for these suggestions via private communication.}, we present here a detailed proof adapted to the maps $U$ and $V$.
	
	\begin{lem}\label{lem::universally-separable}
		Let $\X$ be a non-empty set, let $\Sigma_\Z$ be a $\sigma$--algebra on $\Z=\X\times\{0,1\}$ with $\Pot_{\fin}(\Z)\subseteq\Sigma_\Z$, let $\DDD$ be a set of distributions on $(\Z,\Sigma_\Z)$ and let $\emptyset\neq\HH\subseteq\{0,1\}^\X$ be a hypothesis space with $\Gamma(h)\in\Sigma_\Z$ for any $h\in\HH$. If $\HH$ is universally separable, then $\HH$ is well-behaved with respect to $\DDD$.
	\end{lem}
	\begin{proof}
		Let $\HH_0\subseteq\HH$ be a countable subset fulfilling the condition from Definition~\ref{defn::universally-separable}. Let $m\in\NN$ and $\DD\in\DDD$. We show that the maps $U=U(\HH,m,\DD)$ and $V=V(\HH,m)$ from Definition~\ref{defn::well-behaved} coincide with the maps $U_0=U(\HH_0,m,\DD)$ and $V_0=V(\HH_0,m)$, respectively. As $\HH_0$ is countable, our arguments in Remark~\ref{rem::sup-countability} then imply the well-behavedness of $\HH$. Let $\ul{z},\ul{z}'\in\Z^m$. As $\HH_0$ is a subset of $\HH$, the inequalities $U_0(\ul{z})\leq U(\ul{z})$ and $V_0(\ul{z},\ul{z}')\leq V(\ul{z},\ul{z}')$ are obvious. In order to verify the inequalities $U_0(\ul{z})\geq U(\ul{z})$ and $V_0(\ul{z},\ul{z}')\geq V(\ul{z},\ul{z}')$, it suffices to show that for any $h\in\HH$ and any sequence $\{h_n\}_{n\in\NN}\subseteq\HH_0$ converging pointwise to $h$, the sequence $\{\hat{\er}_\square(h_n)\}_{n\in\NN}$ converges pointwise to $\hat{\er}_\square(h)$ and the sequence $\{\er_\DD(h_n)\}_{n\in\NN}$ converges to $\er_\DD(h)$. Let $h\in\HH$ and let $\{h_n\}_{n\in\NN}\subseteq\HH_0$ be such that for any $x\in\X$ there is $n_x\in\NN$ such that $h(x)=h_n(x)$ for any $n\geq n_x$. Writing $\ul{z}=((x_1,y_1),\dots,(x_m,y_m))$, there exists $n_{\ul{z}}\in\NN$ such that $h(x_i)=h_n(x_i)$ for any $i\in[m]$ and any $n\geq n_{\ul{z}}$. This implies $\hat{\er}_{\ul{z}}(h)=\hat{\er}_{\ul{z}}(h_n)$ for $n\geq n_{\ul{z}}$. Applying these arguments also to $\ul{z}'$, yields
		$$\big|\hspace{0.25pt}\hat{\er}_{\ul{z}'}(h)-\hat{\er}_{\ul{z}}(h)\big|=\big|\hspace{0.25pt}\hat{\er}_{\ul{z}'}(h_n)-\hat{\er}_{\ul{z}}(h_n)\big|$$
		for large enough $n\in\NN$. Hence, we obtain $V(\ul{z},\ul{z}')=V_0(\ul{z},\ul{z}')$. Given $x\in\X$, set $n_x:=\min\{n\in\NN\mid h(x)=h_\ell(x)\text{ for any }\ell\geq n\}$. Further, set $\X_k:=\{x\in\X\mid n_x=k\}$ for $k\in\NN$. Since for any $z=(x,y)\in\Z$ and any $k>1$ we have
		\begin{align*}
			x\in\X_k\ \Leftrightarrow\,&\ k=\min\{n\in\NN\mid h(x)=h_\ell(x)\text{ for any }\ell\geq n\} \\
			\Leftrightarrow\,&\ h(x)\neq h_{k-1}(x) \text{ and } h(x)=h_\ell(x)\text{ for any }\ell\geq k \\
			\Leftrightarrow\,&\ (x,y)\in (\Gamma(h)\triangle\Gamma(h_{k-1})) \text{ and}\\
			\,&\ (x,y)\notin (\Gamma(h)\triangle\Gamma(h_{\ell}))\text{ for any }\ell\geq k,
		\end{align*}
		we can write
		$$\X_k\times\{0,1\}=(\underbrace{\Gamma(h)\triangle\Gamma(h_{k-1})}_{\in\Sigma_\Z})\setminus \bigcup\limits_{\ell\geq k}(\underbrace{\Gamma(h)\triangle\Gamma(h_{\ell})}_{\in\Sigma_\Z})\in\Sigma_\Z,$$
		where $A\triangle B$ denotes the symmetric difference of sets $A,B$. Similar arguments show 
		$$\X_1\times\{0,1\}=\Z\setminus\bigcup\limits_{n\in\NN}(\Gamma(h)\triangle\Gamma(h_n)).$$
		Hence, we obtain
		$1=\DD(\Z)=\sum\limits_{k\in\NN}\DD(\X_k\times\{0,1\})$, as $\Z=\dot{\bigcup\limits_{k\in\NN}}(\X_k\times\{0,1\})$. Therefore, for any $\varepsilon>0$ there exists $\ell\in\NN$ such that
		$$\sum\limits_{k>\ell}\DD(\X_k\times\{0,1\})=\DD\left(\dot{\bigcup\limits_{k>\ell}}(\X_k\times\{0,1\})\right)<\varepsilon,$$
		which implies $\DD(\Gamma(h)\triangle\Gamma(h_n))<\varepsilon$ for any $n> \ell$. Since easy computations show
		$$\big|\hspace{-1.25pt}\er_\DD(h_n)-\er_\DD(h)\big|=\big|\DD(\Gamma(h))-\DD(\Gamma(h_n))\big|\leq \DD(\Gamma(h)\triangle\Gamma(h_n)),$$
		we obtain that the sequence $\{\er_\DD(h_n)\}_{n\in\NN}$ converges to $\er_\DD(h)$. Together with the arguments from above, this yields $U(\ul{z})=U_0(\ul{z})$.
	\end{proof}
	
	The universal separability condition applies to many standard examples of hypothesis spaces (cf.\ Pollard~\cite[Chapter~II, Problems 4, 5 and 7]{Pollard}). However, universal separability is no \emph{necessary} condition for ensuring well-behavedness. In fact, Example~\ref{exa::UCP-measurability-2} presents a quite simple hypothesis space that is not universally separable, but well-behaved with respect to certain sets of distributions.
	
	Further considerations on the measurability of suprema over uncountable families of measurable functions can be found in Dudley~\cite{Dudley1978}, \cite[Chapter~10]{Dudley1984}, Gaenssler~\cite[Chapter~4]{Gaenssler} and Pollard~\cite[Appendix~C]{Pollard}. For instance, \cite[Appendix~C]{Pollard} introduces the concept of \emph{permissible} function classes, which Haussler~\cite[\S\,9.2]{Haussler1992} and Lee~\cite[\S\,2.3]{Lee1996} also take up, e.g.\ to derive similar bounds as in Theorem~\ref{thm::VC-UCP-help}. Recall that the arguments in \cite[Appendix~C]{Pollard}, originally developed for permissible function classes, form the core for our proof of Theorem~\ref{thm::o-minimal-PAC}. 
	Most hypothesis spaces arising in applications can in fact be shown to be permissible (\cite[page~137]{Haussler1992}). The considerations in \cite[Appendix~C]{Pollard} further imply that permissibility constitutes a sufficient condition for well-behavedness. A permissible hypothesis space must be indexed by an analytic subset of a compact metric space, whereas well-behavedness reflects the minimal measure-theoretic requirements of the Fundamental Theorem of Statistical Learning, and is therefore maximally general. In theory, there could be non-permissible hypothesis spaces for which the more general framework of well-behavedness is nonetheless applicable.
	
	\begin{rem}\label{rem::discrete-distributions-measurability}
		The considered set $\DDD$ of distributions also constitutes an important factor that has to be taken into account in the context of measurability. If all distributions in $\DDD$ are discrete, then the Fundamental Theorem of Statistical Learning also applies to hypothesis spaces that are not well-behaved with respect to $\DDD$:
		\begin{enumerate}[(a)]
			\item As already indicated in Remark~\ref{rem::proof-fund-thm}, the measurability of the map $U$ is primarily necessary for the uniform convergence property to be well-defined. However, similar to our definition of PAC learning (see Definition~\ref{defn::PAC}), we can modify our definition of the uniform convergence property. More precisely, we can omit the measurability assumption on $U$ and replace the condition in Definition~\ref{defn::UCP} by the following refinement:
			\\\vspace{-0.2cm}
			
			\begingroup
			\leftskip3.2em
			\rightskip\leftskip
			For any $\varepsilon,\delta\in(0,1)$ there exists $m_0=m_0(\varepsilon,\delta)\in\NN$ such that for any $m\geq m_0$ and any $\DD\in\DDD$ there exists a set $C=C(\varepsilon,\delta,m,\DD)\in\Sigma_\Z^m$ such that
			\begin{align*}
				&C\subseteq \left\{\ul{z}\in\Z^m\left|\ \sup\limits_{h\in\HH}|\er_\DD(h)-\hat{\er}_{\ul{z}}(h)|\leq\varepsilon\right.\right\}\\
				&\text{and }\DD^m(C)\geq 1-\delta.
			\end{align*}
			\endgroup
			\item For the proofs of Theorem~\ref{thm::VC-UCP-help} and, building on it, Theorem~\ref{thm::VC-UCP} both $U$ and $V$ have to measurable, as a key step is to bound their expected values. To adjust these proofs for arbitrary hypothesis spaces, we first note that the maps $U$ and $V$ are certainly $\Pot(\Z^m)$--measurable and $\Pot(\Z^{2m})$--measurable, respectively. If $\DD$ is discrete, then we can write
			$$\DD=\sum\limits_{z\in C_\DD}p_z\delta_z$$
			for suitable $p_z\in[0,1]$ and a finite set $C_\DD\subseteq\Z$, and we can extend\footnote{It is easily verified that the restriction of $\hat{\DD}$ to $\Sigma_\Z$ coincides with $\DD$.} $\DD$ to a distribution $\hat{\DD}$ on $\Pot(\Z)$ by setting 
			$$\hat{\DD}(C):=\DD(\underbrace{C\cap C_\DD}_{\in\Sigma_\Z})$$
			for $C\subseteq\Z$. In our computations, we can then replace the probability space $(\Z,\Sigma_\Z,\DD)$ and its powers $(\Z^k,\Sigma_\Z^k,\DD^k)$ by the ``discretized'' probability space $(\Z,\Pot(\Z),\hat{\DD})$ and its powers $(\Z^k,\Pot(\Z^k),\hat{\DD}^k)$, in order to omit measurability difficulties. With these replacements, following the proofs of Theorem~\ref{thm::VC-UCP-help} and Theorem~\ref{thm::VC-UCP}, we then obtain
			\begin{align*}
				\DD^m(C)&=\hat{\DD}^m(\{\ul{z}\in\Z^m\mid U(\ul{z})\leq\varepsilon\}) \\
				&\geq\hat{\DD}^m(\{\ul{z}\in\Z^m\mid U(\ul{z})\leq\varepsilon(m,\delta)\}) \\
				&\geq 1-\delta
			\end{align*}
			for the finite set $C=\{\ul{z}\in C_\DD^m\mid U(\ul{z})\leq\varepsilon(m,\delta)\}\in\Sigma_\Z$ and any $m\geq m_0$, where
			$$m_0=m_0(\varepsilon,\delta)=\lceil\max\{m_0^{(1)}, m_0^{(2)},m_0^{(3)}\}\rceil.$$
			\item For the uniform convergence property to be well-defined, the proof of Theorem~\ref{thm::UCP-NMSE} requires $U$ to be measurable. However, the modification described in (a) can be incorporated into the proof. In fact, the core of the proof is to justify the inclusion  
			\begin{align*}
				&\left\{\ul{z}\in\Z^m\left|\ \sup\limits_{h\in\HH}\big|\hspace{-1.25pt}\er_\DD(h)-\hat{\er}_{\ul{z}}(h)\big|\leq\frac{\varepsilon}{4}\right.\right\} \\
				\subseteq\; &\left\{\ul{z}\in\Z^m\mid \er_\DD(\A(\ul{z}))-\opt_\DD(\HH)\leq \varepsilon\right\}\!.
			\end{align*}
			for large enough $m\in\NN$, where $\A$ is a learning function that is NMER. In particular, this inclusion ensures that any measurable subset $C\in\Sigma_\Z^m$ (with $\DD^m(C)\geq 1-\delta$) of the former set is also a subset of the latter set.
		\end{enumerate}
		\vspace{-1em}
	\end{rem}
	
	\subsection{Cells}
	
	In this section we mostly follow van den Driess~\cite[Chapter~3, \S\,2]{Van-den-Dries}. Let $\LL$ be a language expanding $\LL_{<}$ and let $\R$ be an o-minimal $\LL$--structure. We endow the domain $R$ with the order-topology $\tau$. Given $m\in\NN$ and a set $X\subseteq R^m$, we endow $R^m$ with the product topology $\tau^m$ and $X$ with the subset topology $\tau_X=\{\mathcal{O}\cap X\mid \mathcal{O}\in\tau^m\}$. Recall that $(R^m,\tau^m)$ is a Hausdorff topological space. We denote by $\B(R^m)$ the Borel $\sigma$--algebra on $R^m$, which is the smallest $\sigma$--algebra containing all open sets in $R^m$. Further, we denote by $\B(X)$ the trace $\sigma$--algebra given by $\B(X)=\{B\cap X\mid B\in\B(R^m)\}$. Recall that $\B(X)$ is the smallest $\sigma$--algebra containing $\tau_X$, and $\B(X)=\{B\in\B(R^m)\mid B\subseteq X\}\subseteq\B(R^m)$ if and only if $X\in\B(R^m)$ (see Bogachev~\cite[Lemma~6.2.4]{Bogachev2}). A map $f\colon X\to R$ is called definable if its graph $\Gamma(f)=\{(\ul{x},f(\ul{x}))\mid \ul{x}\in X\}\subseteq R^{m+1}$ is definable, and it is called continuous if the preimage of any open set $\mathcal{O}\in\tau$ is contained in $\tau_X$. If $X$ is definable, then we set
	\begin{align*}
		C(X)&:=\{f\colon X\to R\mid f\text{ is definable and continuous}\}, \\
		C_\infty(X)&:=C(X)\cup\{-\infty,\infty\},
	\end{align*}
	where we regard $-\infty$ and $\infty$ as constant functions on $X$. For $f,g\in C_\infty(X)$ we write $f<g$ to indicate that $f(\ul{x})<g(\ul{x})$ for all $\ul{x}\in X$, and in this case we set
	$$(f,g)_X:=\{(\ul{x},r)\in X\times R\mid f(\ul{x})<r<g(\ul{x})\}.$$
	Note that $(f,g)_X$ is a definable subset of $R^{m+1}$.
	
	\begin{defn}\label{defn::cell}
		Let $m\in\NN$ and let $(i_1,\dots,i_m)\in\{0,1\}^m$. An \textbf{$(i_1,\dots,i_m)$--cell} is a definable subset of $R^m$ obtained by induction on $m$ as follows:
		\begin{enumerate}[(i)]
			\item A $(0)$--cell is a singleton $\{r\}\subseteq R$ (a ``point''), a $(1)$--cell is an interval $(a,b)\subseteq R$ with $a,b\in R$.
			\item Supposing that $(i_1,\dots,i_m)$--cells are already defined, an $(i_1,\dots,i_m,0)$--cell is the graph $\Gamma(f)$ of a function $f\in C(X)$, where the set $X$ is an $(i_1,\dots,i_m)$--cell; further, an $(i_1,\dots,i_m,1)$--cell is a set $(f,g)_X$, where $X$ is an $(i_1,\dots,i_m)$--cell and $f,g\in C_\infty(X)$ with $f<g$. 
		\end{enumerate}
		A set $X\subseteq R^m$ is called a \textbf{cell} if it is an $(i_1,\dots,i_m)$--cell for some tuple $(i_1,\dots,i_m)\in\{0,1\}^m$. 
	\end{defn}
	
	\begin{lem}\label{lem::cells-Borel}
		Let $m\in\NN$, let $(i_1,\dots,i_m)\in\{0,1\}^m$ and let $X$ be an $(i_1,\dots,i_m)$--cell. Then $X\in\B(R^m)$. 
	\end{lem}
	\begin{proof}
		We prove the statement by induction on $m$. Any $(0)$--cell is a singleton $\{r\}\subseteq R$. As $(R,\tau)$ is a Hausdorff topological space, we obtain $\{r\}\in\B(R)$. Any $(1)$--cell is an interval $(a,b)\subseteq R$ with $a,b\in R$, and by definition of the order-topology we obtain $(a,b)\in\tau\subseteq\B(R)$. In the following let $X$ be an $(i_1,\dots,i_m)$--cell, and assume that $X\in\B(R^m)$. Further, consider $f\in C(X)$. Recall that, as a continuous function, $f$ is $\B(X)$--$\B(R)$--measurable\footnote{Given two measurable spaces $(\Omega_1,\Sigma_1), (\Omega_2,\Sigma_2)$, a function $f\colon\Omega_1\to\Omega_2$ is called \textbf{$\Sigma_1$--$\Sigma_2$--measurable} if $f^{-1}(A)\in\Sigma_1$ for any $A\in\Sigma_2$.} (see Bogachev~\cite[Lemma~6.2.2]{Bogachev2}). In particular, the functions 
		\begin{align*}
			g_f&\colon X\times R\to R,\ (\ul{x},r)\mapsto f(\ul{x}), \\
			\pi_2&\colon X\times R\to R,\ (\ul{x},r)\mapsto r
		\end{align*}
		are $(\B(X)\otimes\B(R)$)--$\B(R)$--measurable. Indeed, for any set $B\in\B(R)$ we have $g_f^{-1}(B)=f^{-1}(B)\times R\in\B(X)\otimes\B(R)$ and $\pi_2^{-1}(B)=X\times B\in\B(\X)\otimes\B(R)$. Thus, also the function 
		$$T_f=g_f-\pi_2\colon X\times R,\ (\ul{x},r)\mapsto f(\ul{x})-r$$ 
		is $\B(X)\otimes\B(R)$--$\B(R)$--measurable. From the assumption $X\in\B(R^m)$ it follows that $\B(X)\subseteq\B(R^m)$, which implies the inclusions $\B(X)\otimes\B(R)\subseteq\B(R^m)\otimes\B(R)\subseteq\B(R^{m+1})$. Hence, the $(i_1,\dots,i_m,0)$--cell $\Gamma(f)=T_f^{-1}(\{0\})$ is a member of $\B(R^{m+1})$. Moreover, we obtain 
		$$T_f^{-1}((-\infty,0)), T_f^{-1}((0,\infty))\in\B(R^{m+1}).$$
		Considering $g,h\in C(X)$ with $g<h$, this implies that the $(i_1,\dots,i_m,1)$--cell 
		\begin{align*}
			(g,h)_X&=\{(\ul{x},r)\in X\times R\mid g(\ul{x})<r<h(\ul{x})\} \\
			&=\{(\ul{x},r)\in X\times R\mid g(\ul{x})<r\}\cap \{(\ul{x},r)\in X\times R\mid r<h(\ul{x})\} \\
			&=T_g^{-1}((-\infty,0))\cap T_h^{-1}((0,\infty))
		\end{align*}
		is a member of $\B(R^{m+1})$. On the other hand, if $f\in\{-\infty,+\infty\}$, then the sets $\{(\ul{x},r)\in X\times R\mid f(\ul{x})<r\}$ and $\{(\ul{x},r)\in X\times R\mid r<f(\ul{x})\}$ are either empty or coincide with the set $X\times R$, respectively, and hence they are members of $\B(R^{m+1})$. This shows that, given $g,h\in C_\infty(X)$ with $g<h$, the $(i_1,\dots,i_m,1)$--cell $(g,h)_X$ is a member of $\B(R^{m+1})$, completing our inductive argumentation.
	\end{proof}
	
	\subsection{Examples}
	
	In this section, we present several examples that stress the importance and necessity of our measurability assumptions and refinements. For the construction of these examples, the existence of non-measurable maps is indispensable. It is well-known that there exist subsets of the reals that are non-Borel. In fact, within the framework of ZFC, one can specify examples of sets that are not Lebesgue measurable and thus not Borel (cf.\ Bogachev~\cite[Example~1.7.7]{Bogachev}). The following result can be used to construct further examples of non-measurable sets in product spaces.
	
	\begin{lem}\label{lem::product-not-measurable}
		Let $(\Omega_1,\Sigma_1)$ and $(\Omega_2,\Sigma_2)$ be measurable spaces. Further, let $A_1\in\Omega_1$ with $A_1\notin\Sigma_1$ and $A_2\in\Omega_2$ with $A_2\notin\Sigma_2$. Then $A_1\times B_2\notin \Sigma_1\otimes\Sigma_2$ and $B_1\times A_2\notin \Sigma_1\otimes\Sigma_2$ for any $\emptyset\neq B_1\subseteq \Omega_1$ and any $\emptyset\neq B_2\subseteq \Omega_2$.
	\end{lem}
	\begin{proof}
		We only prove $A_1\times B_2\notin \Sigma_1\otimes\Sigma_2$ for any $\emptyset\neq B_2\subseteq\Omega_2$. A symmetric argument shows $B_1\times A_2\notin\Sigma_1\otimes\Sigma_2$ for any $\emptyset\neq B_1\subseteq\Omega_1$. Let $\emptyset\neq B_2\subseteq\Omega_2$ and choose an arbitrary $b\in B_2$. Consider the map
		$$f_b\colon \Omega_1\to\Omega_1\times\Omega_2,\ \omega_1\mapsto (\omega_1,b).$$
		The map $f_b$ is $\Sigma_1$--($\Sigma_1\otimes\Sigma_2$)--measurable, since for any $E_1\in\Sigma_1, E_2\in\Sigma_2$ we have
		$$f_b^{-1}(E_1\times E_2)=\left.\begin{cases}
			E_1&\text{if }b\in E_2 \\
			\emptyset&\text{otherwise}
		\end{cases}\right\}\in\Sigma_1,$$
		and $\Sigma_1\otimes\Sigma_2$ is generated by the system $\{E_1\times E_2\mid E_1\in\Sigma_1,E_2\in\Sigma_2\}$. Hence, $A_1\times B_2\in\Sigma_1\otimes\Sigma_2$ would imply $A_1=f_b^{-1}(A_1\times B_2)\in\Sigma_1$, which is a contradiction.
	\end{proof}
	
	\begin{rem}
		In the literature, the condition in Definition~\ref{defn::PAC} usually has the following simpler form: \\\vspace{-0.2cm}
		
		\begingroup
		\leftskip3.2em
		\rightskip\leftskip
		For any $\varepsilon,\delta\in(0,1)$ there exists $m_0=m_0(\varepsilon,\delta)\in\NN$ such that for any $m\geq m_0$ and any $\DD\in\DDD$ the following inequality holds true:
		\begin{align*}
			\DD^m(\{\ul{z}\in\Z^m\mid \er_\DD(\A(\ul{z}))-\opt_\DD(\HH)\leq\varepsilon\})\geq 1-\delta.
		\end{align*}
		\endgroup
		However, this condition is in general not well-defined, as the set 
		$$\{\ul{z}\in\Z^m\mid \er_\DD(\A(\ul{z}))-\opt_\DD(\HH)\leq\varepsilon\}$$ 
		is not necessarily measurable (see Example~\ref{exa::PAC-measurability}). Therefore, we ask for the existence of a set $C=C(\varepsilon,\delta,m,\DD)\in\Sigma_\Z^m$ with $\DD^m(C)\geq 1-\delta$ that is contained in the above set, which itself might be not a member of $\Sigma_\Z^m$. This measurability refinement is inspired by \cite[\S\,2]{Blumer}. 
	\end{rem}	
	
	\begin{exa}\label{exa::PAC-measurability}
		Consider $\X=\RR$ and let $A\subseteq\RR$ be such that $A\notin\B(\RR)$. Set $\Z=\RR\times\{0,1\}$, $\Sigma_\Z=\B(\RR)\otimes\Pot(\{0,1\})$ and consider $C=A\times\{0,1\}$. Inductively applying Lemma~\ref{lem::product-not-measurable} yields $C^m\notin\Sigma_\Z^m$ for any $m\in\NN$. Further, consider the hypothesis space $\HH=\{\1_\emptyset, \1_{\{0\}}\}$ and the learning function $\A$ for $\HH$ given by 
		$$\A(\ul{z})=\begin{cases}
			\1_\emptyset&\text{if }\ul{z}\in C^m,  \\
			\1_{\{0\}}&\text{otherwise},
		\end{cases}$$
		for any $m\in\NN$ and $\ul{z}\in\Z^m$, and set $\DD=\delta_{(0,0)}$. Clearly, we have $\Gamma(h)\in\Sigma_\Z$ for any $h\in\HH$, and $\opt_\DD(\HH)=\er_\DD(\1_\emptyset)=0$ as well as $\er_\DD(\1_{\{0\}})=1$, which implies
		\begin{align*}
			&\;\{\ul{z}\in\Z^m\mid \er_\DD(\A(\ul{z}))-\opt_\DD(\HH)\leq\varepsilon\} \\
			=&\;\{\ul{z}\in\Z^m\mid \er_\DD(\A(\ul{z}))\leq\varepsilon\} \\
			=&\;\{\ul{z}\in\Z^m\mid \er_\DD(\A(\ul{z}))=0\} \\
			=&\;\{\ul{z}\in\Z^m\mid \A(\ul{z})=\1_\emptyset\} \\
			=&\;C^m\notin\Sigma_\Z^m
		\end{align*}
		for any $\varepsilon\in(0,1)$ and any $m\in\NN$. 
	\end{exa}
	
	The next example shows that there exist hypothesis spaces that are not well-behaved. Moreover, we address the question whether the Fundamental Theorem of Statistical Learning still applies to such hypothesis spaces. 
	
	\begin{exa}\label{exa::UCP-measurability}
		Consider $\X=\RR$ and let $A\subseteq\RR$ be such that $A\notin\B(\RR)$. Set $\Z=\RR\times\{0,1\}$, $\Sigma_\Z=\B(\RR)\otimes\Pot(\{0,1\})$ and consider $C=A\times\{1\}\subseteq\Z$. Inductively applying Lemma~\ref{lem::product-not-measurable} yields $C^m\notin\Sigma_\Z^m$ for any $m\in\NN$. Further, consider the hypothesis space $$\HH=\{\1_\emptyset\}\cup\left\{\left.\1_{\RR\setminus\{w\}}\ \right|\, w\in \RR\setminus A\right\}\!,$$ 
		and set $\DD=\delta_{(a_0,1)}$ for some fixed $a_0\in A$. Clearly, we have  $\Gamma(h)\in\Sigma_\Z$ for any $h\in\HH$, and $\er_\DD(\1_\emptyset)=1$ as well as $\er_\DD(h_w)=0$ for any $w\in\RR\setminus A$. Next, we consider the map $U=U(\HH,m,\DD)$ from Definition~\ref{defn::UCP} and compute the value $U(\ul{z})$ for $\ul{z}\in\Z^m$. If $\ul{z}=(z_1,\dots,z_m)\in C^m$, i.e.\ $z_i=(a_i,1)$ with $a_i\in A$ for any $i\in[m]$, then $U(\ul{z})=0$. In fact, we compute
		$$\big|\hspace{-0.3pt}\er_\DD(\1_\emptyset)-\hat{\er}_{\ul{z}}(\1_\emptyset)\hspace{0.3pt}\big|=|1-1|=0,$$
		since $\1_\emptyset(a_i)=0\neq 1$ implies $z_i\notin\Gamma(\1_\emptyset)$ for any $i\in[m]$. Moreover, for any $w\in\RR\setminus A$ we obtain
		$$\big|\hspace{-0.3pt}\er_\DD(\1_{\RR\setminus\{w\}})-\hat{\er}_{\ul{z}}(\1_{\RR\setminus\{w\}})\hspace{0.3pt}\big|=|0-0|=0,$$
		since $\1_{\RR\setminus\{w\}}(a_i)=0$ implies $z_i\in\Gamma(\1_{\RR\setminus\{w\}})$ for any $i\in[m]$. On the other hand, if $\ul{z}=(z_1,\dots,z_m)\notin C^m$, then $z_j=(x_j,y_j)\notin C$ for some $j\in[m]$. We now distinguish two cases. If $y_j=0$, then we obtain
		$$\hat{\er}_{\ul{z}}(\1_\emptyset)=\frac{1}{m}\sum\limits_{i\in[m]} \1_{\Z\setminus\Gamma(\1_\emptyset)}(z_i)=\frac{1}{m}\sum\limits_{j\neq i\in[m]} \1_{\Z\setminus\Gamma(\1_\emptyset)}(z_i)<1=\er_\DD(\1_\emptyset),$$
		since $h(x_j)=0=y_j$ implies $z_j\in\Gamma(\1_\emptyset)$. On the other hand, if $y_j=1$, then $x_j\notin A$. Hence, the hypothesis $h_j=\1_{\RR\setminus\{x_j\}}$ is a member of $\HH$, and we obtain 
		\begin{align*}
			\hat{\er}_{\ul{z}}(h_j)&=\frac{1}{m}\sum\limits_{i\in[m]} \1_{\Z\setminus\Gamma(h_j)}(z_i)\\
			&=\frac{1}{m}+\frac{1}{m}\sum\limits_{j\neq i\in[m]} \1_{\Z\setminus\Gamma(h_j)}(z_i)>0=\er_\DD(h_j),
		\end{align*}
		since $h_j(x_j)=\1_{\RR\setminus\{x_j\}}(x_j)=0\neq y_j$ implies $z_j\notin\Gamma(h_j)$. 
		Combining everything, we obtain $U(\ul{z})=0$ if and only if $\ul{z}\in C^m$, which implies
		$$U^{-1}(\{0\})=C^m\notin \Sigma_\Z^m.$$
		Thus, $U$ is not $\Sigma_\Z^m$--measurable. In particular, the hypothesis space $\HH$ is not well-behaved with respect to $\DDD=\{\delta_{(a_0,1)}\}$. It is easily verified that $\mathrm{vc}(\HH)=1$. Therefore, the question arises whether $\HH$ is PAC learnable with respect to $\DDD=\{\delta_{(a_0,1)}\}$, even though it does not meet the measurability requirements of the Fundamental Theorem of Statistical Learning. In fact, the answer is an affirmative one. The reason for this is that the distribution $\delta_{(a_0,1)}$ is discrete. In fact, if all distributions in $\DDD$ are discrete, the results in Section~\ref{sec::fundthm} also apply to hypothesis spaces that are not well-behaved with respect to $\DDD$. This exception is further explained in Remark~\ref{rem::discrete-distributions-measurability}.
	\end{exa}
	
	\begin{exa}\label{exa::complete}
		Consider $\X=\RR$ and set  $\Z=\RR\times\{0,1\}$. As $\B(\RR^2)=\B(\RR)\otimes\B(\RR)$ and $\B(\{0,1\})=\Pot(\{0,1\})$, applying Lemma~\ref{lem::product-trace} yields $\B(\Z)=\B(\RR)\otimes\Pot(\{0,1\})$. Let $\U(0,1)$ denote the standard uniform distribution defined on $\B(\RR)$, let $\delta_1$ denote the Dirac measure on $\{0,1\}$, and consider the distribution $\DD=\U(0,1)\otimes \delta_1$ on $\B(\Z)$. Then for any $m\in\NN$ the probability space $(\Z^m,\B(\Z)^m,\DD^m)$ is not complete. Indeed, for any set $A\subseteq\RR$ with $A\notin\B(\RR)$, inductively applying 
		Lemma~\ref{lem::product-not-measurable} yields $(A\times\{0\})^m\notin\B(\Z)^m$. Additionally, we have $(A\times\{0\})^m\subseteq(\RR\times\{0\})^m$, and we compute $$\DD^m((\RR\times\{0\})^m)=\DD(\RR\times\{0\})^m=(\U(0,1)(\RR)\cdot\underbrace{\delta_1(\{0\})}_{=0})^m=0.$$
		Thus, the distribution set $\DDD=\{\DD\}$ does not satisfy the completeness condition in Theorem~\ref{thm::o-minimal-PAC}. Similar arguments apply to any distribution of the form $\DD=\PP\otimes\delta_y$, where $\PP$ is any distribution defined on $\B(\RR)$ and $\delta_{y}$ is the Dirac measure for a given $y\in\{0,1\}$.
	\end{exa}
	
	\begin{exa}\label{exa::UCP-measurability-2}
		Set $\X=\RR$, $\Z=\RR\times\{0,1\}$ and $\Sigma_\Z=\B(\RR)\otimes\Pot(\{0,1\})$. As in Example~\ref{exa::complete}, one can show that $\Sigma_\Z=\B(\Z)$. In Example~\ref{exa::UCP-measurability} we showed that the hypothesis space $$\{\1_\emptyset\}\cup\left\{\left.\1_{\RR\setminus\{w\}}\ \right|\, w\in \RR\setminus A\right\}$$ is not well-behaved with respect to a very simple distribution set by exploiting that the set $A\subseteq\RR$ is not a member of $\B(\RR)$. In a certain way, the hypothesis space has inherited the unfavorable property of being non-measurable from the set $A$. We now instead consider the simpler hypothesis space 
		$$\HH=\left\{\left.\1_{\RR\setminus\{w\}}\ \right|\, w\in \RR\right\}\!$$ 
		(cf.\ also Blumer, Ehrenfeucht, Haussler and Warmuth~\cite[page~954]{Blumer}). First, we notice that $\HH=\HH^\varphi=\{\1_{\varphi(\RR;w)}\mid w\in\RR\}$ for the formula $\varphi(x;p)$ given by $x\neq p$. We aim at verifying that $\HH$ is well-behaved with respect to $\DDD^\ast$. \linebreak 
		There are two natural approaches we could take. On the one hand, we could show that $\HH$ is universally separable and then apply Lemma~\ref{lem::universally-separable} to deduce its well-behavedness. However, it can be easily verified that $\HH$ is not universally separable. On the other hand, we could check whether Theorem~\ref{thm::o-minimal-PAC} is applicable. This is also not the case, since the completeness condition in Theorem~\ref{thm::o-minimal-PAC} is not satisfied in this example. For instance, one can consider the distribution $\DD=\U(0,1)\otimes\delta_1$, for which we have disproven the completeness condition in Example~\ref{exa::complete}. For these reasons, we directly prove the measurability of the maps $U$ and $V$ from Definition~\ref{defn::well-behaved}. To this end, we first consider the map $U=U(\HH,m,\DD)$ for $m\in\NN$ and $\DD\in\DDD^\ast$ fulfilling the condition 
		\begin{align}\label{eqn::cond0}
			\er_\DD(h)=0 \text{ for any }h\in\HH. 
		\end{align}
		For instance, the distribution $\DD=\U(0,1)\otimes\delta_1$ satisfies condition~\eqref{eqn::cond0}, since straightforward computations show
		$$\er_\DD\big(\1_{\RR\setminus\{w\}}\big)=\DD\left(\Z\setminus\Gamma\big(\1_{\RR\setminus\{w\}}\big)\right)=\U(0,1)(\{w\})=0$$
		for any $w\in\RR$. In general, condition~\eqref{eqn::cond0} implies that the map $U$ is given by
		$$U\colon\Z^m\to\left.\left\{\tfrac{k}{m}\ \right|\, k\in[m]_0\right\},\ \ul{z}\mapsto \sup\limits_{w\in\RR}\hat{\er}_{\ul{z}}\big(h_w\big),$$
		writing $h_w$ for the hypothesis $\1_{\RR\setminus\{w\}}\in\HH$. For any $k\in[m]_0$ we compute 
		\begin{align*}
			&\left\{\ul{z}\in\Z^m \,\left|\ U(\ul{z})\geq\tfrac{k}{m}\right.\right\} \\
			=&\left\{\ul{z}\in\Z^m\, \left|\ \exists w\in\RR\colon\hat{\er}_{\ul{z}}\big(h_w\big)\geq\tfrac{k}{m}\right.\right\} \\
			=&\bigcup\limits_{\substack{I\subseteq[m]\\ |I|\geq k}}\, \bigcup\limits_{w\in\RR} C_{w,I},
		\end{align*}
		where 
		$$C_{w,I}=\bigtimes\limits_{i\in[m]}\! C_{w,I}^{(i)}$$
		with 
		\begin{align*}
			C_{w,I}^{(i)}&=\begin{cases}
				\Z\setminus\Gamma\big(h_w\big)&\text{if }i\in I, \\
				\Gamma\big(h_w\big)&\text{otherwise},
			\end{cases}
		\end{align*}
		for $x\in\RR$, $I\subseteq[m]$ and $i\in[m]$. The set $C_{w,I}$ can be written as 
		\begin{align*}
			C_{w,I}=\{((x_1,y_1),\dots,(x_1,y_m))\in\Z^m\mid\forall i\in[m]\colon y_i=h_w(x_i)\Leftrightarrow i\not\in I\}.
		\end{align*}
		Set $\Z_{\ul{y}}^m=\RR\times\{y_1\}\times\dots\times\RR\times\{y_m\}$ for $\ul{y}=(y_1,\dots,y_m)\in\{0,1\}^m$. The finite union of these sets coincides with $\Z^m$, i.e.\ we have
		$$\Z^m=\bigcup\limits_{\ul{y}\in\{0,1\}^m}\Z_{\ul{y}}^m.$$
		Thus, for $I\subseteq [m]$ we can write
		\begin{align*}
			&\bigcup\limits_{w\in\RR} C_{w,I} \\
			=&\left\{\ul{z}\in\Z^m\mid \exists w\in\RR\,\forall i\in[m]\colon y_i=h_w(x_i)\Leftrightarrow i\not\in I \right\} \\
			=&\bigcup\limits_{\ul{y}\in\{0,1\}^m} \left\{\left.\ul{z}\in\Z_{\ul{y}}^m\ \right|\,  \exists w\in\RR\,\forall i\in[m]\colon y_i=h_w(x_i)\Leftrightarrow i\not\in I \right\} \\
			=&\bigcup\limits_{\ul{y}\in\{0,1\}^m} \left\{\left.\ul{z}\in\Z_{\ul{y}}^m\ \right|\, \ul{z}\text{ satisfies all conditions in }\mathcal{R}\big(I,\ul{y}\:\!\big)\right\} \\
			=&\bigcup\limits_{\ul{y}\in\{0,1\}^m} \bigcap\limits_{R\in\mathcal{R}(I,\ul{y})} \left\{\left.\ul{z}\in\Z_{\ul{y}}^m\ \right|\, \ul{z}\text{ satisfies condition }R\right\}\!, 
		\end{align*}
		where the finite set $\mathcal{R}\big(I,\ul{y}\:\!\big)$ of conditions is given by 
		$$\mathcal{R}\big(I,\ul{y}\:\!\big)\!=\!\left.\begin{cases}
			x_i= x_j&\text{ for any }i,j\in I\text{ with }\{y_i,y_j\}=\{1\} \\ 
			x_i= x_j&\text{ for any }i,j\in [m]\setminus I\text{ with }\{y_i,y_j\}=\{0\} \\
			x_i= x_j&\text{ for any }i\in I,j\in [m]\setminus I\text{ with } y_i=1\text{ and }y_j=0  \\
			x_i\neq x_j&\text{ for any }i,j\in I\text{ with }\{y_i,y_j\}=\{0,1\} \\ 
			x_i\neq x_j&\text{ for any }i,j\in [m]\setminus I\text{ with }\{y_i,y_j\}=\{0,1\} \\ 
			x_i\neq x_j&\text{ for any }i\in I,j\in [m]\setminus I\text{ with }\{y_i,y_j\}=\{0\} \\ 
			x_i\neq x_j&\text{ for any }i\in I,j\in [m]\setminus I\text{ with }\{y_i,y_j\}=\{1\} 
		\end{cases}\right\}.$$
		If $\mathcal{R}\big(I,\ul{y}\:\!\big)=\emptyset$, then the intersection
		$$\bigcap\limits_{R\in\mathcal{R}(I,\ul{y})} \left\{\left.\ul{z}\in\Z_{\ul{y}}^m\ \right|\, \ul{z}\text{ satisfies condition }R\right\}$$
		is replaced by the set $\Z^m_{\ul{y}}$. Our computations show that the set $$\left\{\ul{z}\in\Z^m\left|\ U(\ul{z})\geq\tfrac{k}{m}\right.\right\}$$ 
		can be written as a finite Boolean combination of the open sets $\Z_{\ul{y}}^m$ and
		$\mathcal{O}_{i,j}=\left\{\left.\ul{z}\in\Z_{\ul{y}}^m\ \right|\, x_i\neq x_j\right\}$, where $i,j\in[m]$ and $\ul{y}\in\{0,1\}^m$. Indeed, we have
		$$\left\{\left.\ul{z}\in\Z^m_{\ul{y}}\ \right|\, x_i=x_j\right\}=\Z_{\ul{y}}^m\setminus \mathcal{O}_{i,j}.$$
		Hence, we obtain $\big\{\ul{z}\in\Z^m\mid U(\ul{z})\geq\tfrac{k}{m}\big\}\in\Sigma_\Z^m$. \\
		The above arguments can be extended to show that the map $U=U(\HH,m,\DD)$ is $\Sigma_\Z^m$--measurable if, more generally, the map $\er_\DD$ is constant on $\HH$. Furthermore, similar arguments can be applied to prove that the map $V=V(\HH,m)$ is always $\Sigma_\Z^{2m}$--measurable, regardless of the set of distributions considered. Showing that the map $U=U(\HH,m,\DD)$ is $\Sigma_\Z^m$--measurable for an arbitrary distribution $\DD$ requires more complex arguments. 
	\end{exa}
	
	The above example illustrates that also very simple and benign hypothesis spaces require lengthy and technical computations when it comes to verifying their well-behavedness.
	
	\vfill
	
	{\textbf{Data and Materials Availability:} Not applicable.}
	
	{\textbf{Code Availability:} Not applicable.}
	
	{\textbf{Ethical Approval:} Not applicable.}
	
	{\textbf{Consent to Participate:} Not applicable.}
	
	{\textbf{Consent for Publication:} Not applicable. The paper does not include data or images that require permissions
		to be published.}
	

\begin{thebibliography}{99}
		\addcontentsline{toc}{section}{References}
		
		\bibitem{AADFT} \textsc{N.~Ackerman, J.~Asilis, J.~Di, C.~Freer} and \textsc{J.-B.~Tristan}, `Computable PAC Learning of Continuous Features', \emph{Proceedings of the 37th Annual ACM/IEEE Symposium on Logic in Computer Science} (2022) Article~7, \href{https://doi.org/10.1145/3531130.3533330}{doi:10.1145/3531130.3533330}.
		
		\bibitem{Anthony-Bartlett} \textsc{M.~Anthony} and \textsc{P.~L.~Bartlett}, \textsl{Neural Network Learning: Theoretical Foundations}, (Cambridge University Press, Cambridge, 1999), \href{https://doi.org/10.1017/CBO9780511624216}{doi:10.1017/CBO9780511624216}.
		
		\bibitem{Ben-David-Benedek-Mansour} \textsc{S.~Ben-David, G.~M.~Benedek} and \textsc{Y.~Mansour}, `A Parameterization Scheme for Classifying Models of PAC Learnability', \textsl{Inf.~Comput.\ }\textbf{120} (1995) 11--21, \href{https://doi.org/10.1006/inco.1995.1094}{doi:10.1006/inco.1995.1094}.
		
		\bibitem{Ben-David-Itai-Kushilevitz} \textsc{S.~Ben-David, A.~Itai} and \textsc{E.~Kushilevitz}, `Learning by Distances', \textsl{Inf.~Comput.\ }\textbf{117} (1995) 240–250, \href{https://doi.org/10.1006/inco.1995.1042}{doi:10.1006/inco.1995.1042}.
		
		\bibitem{Blumer} \textsc{A.~Blumer, A.~Ehrenfeucht, D.~Haussler} and \textsc{M.~K.~Warmuth}, `Learnability and the Vapnik--Chervonenkis Dimension', \textsl{J.\ Assoc.\ Comput.\ Mach.\ }\textbf{36} (1989) 929--965, \href{https://doi.org/10.1145/76359.76371}{doi:10.1145/76359.76371}.
		
		\bibitem{Bogachev} \textsc{V.~I.~Bogachev}, \textsl{Measure Theory}, Vol.\ 1 (Springer, Berlin, 2007), \href{https://doi.org/10.1007/978-3-540-34514-5}{doi:10.1007/978-3-540-34514-5}.
		
		\bibitem{Bogachev2} \textsc{V.~I.~Bogachev}, \textsl{Measure Theory}, Vol.\ 2 (Springer, Berlin, 2007), \href{https://doi.org/10.1007/978-3-540-34514-5}{doi:10.1007/978-3-540-34514-5}.
		
		\bibitem{Chase-Freitag} \textsc{H.~Chase} and \textsc{J.~Freitag}, `Model Theory and Machine Learning', \textsl{Bull.\ Symb.\ Log.\ }\textbf{25} (2019) 319–332, \href{https://doi.org/10.1017/bsl.2018.71}{doi:10.1017/bsl.2018.71}.
		
		\bibitem{DM} \textsc{C.~Dellacherie} and \textsc{P.-A.~Meyer}, \textsl{Probabilities and Potential}, North-Holland Math.\ Stud.\ 29 (North-Holland, Amsterdam, 1978), \href{https://doi.org/10.1016/S0304-0208(08)72749-X}{doi:10.1016/S0304-0208(08)72749-X}.
		
		\bibitem{Van-den-Dries} \textsc{L.~van den Dries}, \textsl{Tame Topology and O-minimal Structures}, Lond.\ Math.\ Soc.\ Lect.\ Note Ser.\ 248 (Cambridge University Press, Cambridge, 1998), \href{
			https://doi.org/10.1017/CBO9780511525919}{doi:10.1017/CBO9780511525919}.
		
		\bibitem{Dudley1978} \textsc{R.~M.~Dudley}, `Central Limit Theorems for Empirical Measures', \textsl{Ann.\ Probab.\ }\textbf{6} (1978) 899--929, \href{https://doi.org/10.1214/aop/1176995384}{doi:10.1214/aop/1176995384}.
		
		\bibitem{Dudley1984}  \textsc{R.~M.~Dudley}, `A Course on Empirical Processes', \textsl{École d'Été de Probabilités de Saint-Flour XII -- 1982} (R.~M.~Dudley, H.~Kunita and F.~Ledrappier; ed.\ P.~L.~Hennequin), Lect.\ Notes Math.\ 1097 (Springer, Berlin, 1984) 1–142, \href{https://doi.org/10.1007/BFb0099431}{doi:10.1007/BFb0099431}.
		
		\bibitem{Gaenssler} \textsc{P.~Gaenssler}, \textsl{Empirical Processes}, IMS Lect.\ Notes, Monogr.\ Ser.\ 3 (IMS, Hayward, 1983), \href{https://projecteuclid.org/euclid.lnms/1215465233#toc}{doi:10.1214/lnms/1215465233}.
		
		\bibitem{Haussler1988} \textsc{D.~Haussler}, `Quantifying Inductive Bias:
		AI Learning Algorithms and Valiant's Learning Framework', \textsl{Artif.\ Intell.\ }\textbf{36}
		(1988) 177--221, \href{https://doi.org/10.1016/0004-3702(88)90002-1}{doi:10.1016/0004-3702(88)90002-1}.
		
		\bibitem{Haussler1992} \textsc{D.~Haussler}, `Decision Theoretic Generalizations of the PAC Model for Neural Net and Other Learning Applications', \textsl{Inf.\ Comput.\ }\textbf{100} (1992) 78--150, \href{https://doi.org/10.1016/0890-5401(92)90010-D}{doi:10.1016/0890-5401(92)90010-D}.
		
		\bibitem{Kaiser} \textsc{T.~Kaiser}, `First order tameness of measures', \textsl{Ann.\ Pure Appl.\ Logic} \textbf{163} (2012) 1903--1927, \href{https://doi.org/10.1016/j.apal.2012.06.002}{doi:10.1016/j.apal.2012.06.002}.
		
		\bibitem{Karpinski-Macintyre} \textsc{M.~Karpinski} and \textsc{A.~Macintyre}, `Approximating Volumes and Integrals in o-Minimal and p-Minimal Theories', \textsl{Connections between Model Theory and Algebraic and Analytic Geometry} (ed.\ A.~Macintyre), Quad.\ Mat.\ 6 (Dipartimento di Matematica della Seconda Università di Napoli, Caserta, 2000) 149--177.
		
		\bibitem{Laskowski} \textsc{M.~C.~Laskowski}, `Vapnik--Chervonenkis Classes of Definable Sets', \textsl{J.\ Lond.\ Math.\ Soc., II.\ Ser.\ }\textbf{45} (1992) 377--384, \href{https://doi.org/10.1112/jlms/s2-45.2.377}{doi:10.1112/jlms/s2-45.2.377}.
		
		\bibitem{Lee1996} \textsc{W.~S.~Lee}, \textsl{Agnostic Learning and Single Hidden Layer Neural Networks}, Doctoral Thesis, Australian National University, 1996.
		
		\bibitem{vonLuxburg-Schoelkopf} \textsc{U.~von Luxburg} and \textsc{B.~Schölkopf}, `Statistical Learning Theory: Models, Concepts, and Results', \textsl{Inductive Logic} (Eds.\ D.~M.~Gabbay, S.~Hartmann and J.~Woods), Handbook of the History of Logic, Vol.~10 (Elsevier, Oxford, 2011) 651–706, \href{https://doi.org/10.1016/B978-0-444-52936-7.50016-1}{doi:10.1016/B978-0-444-52936-7.50016-1}.
		
		\bibitem{Marker} \textsc{D.~Marker}, \textsl{Model Theory: An Introduction}, Grad.\ Texts Math.\ 217 (Springer, New York, 2002), \href{https://doi.org/10.1007/b98860}{doi:10.1007/b98860}. 
		
		\bibitem{Mendelson-Smola} \textsc{S.~Mendelson} and \textsc{A.~J.~Smola} (Eds.), \textsl{Advanced Lectures on Machine Learning: Machine Learning Summer School 2002 Canberra, Australia, February 11-22, 2002, Revised Lectures}, Lect.\ Notes Comput.\ Sci.\ 2600 (Springer, Berlin, 2003), \href{https://doi.org/10.1007/3-540-36434-X}{doi:10.1007/3-540-36434-X}. 
		
		\bibitem{Pestov} \textsc{V.~Pestov}, `PAC learnability versus VC dimension: a footnote to a basic result of statistical learning', \textsl{IJCNN 2011 Conference Proceedings} (IEEE, Piscataway, 2011) 1141--1145, \href{https://doi.org/10.1109/IJCNN.2011.6033352}{doi:10.1109/IJCNN.2011.6033352}.
		
		\bibitem{Pillay-Steinhorn} 
		\textsc{A.~Pillay} and \textsc{C.~Steinhorn}, 
		`Definable sets in ordered structures', 
		I, \textsl{Trans.\ Amer.\ Math.\ Soc.} \textbf{295} (1986) 565--592,  \href{https://doi.org/10.1090/S0002-9947-1986-0833697-X}{doi:10.1090/S0002-9947-1986-0833697-X}.
		
		\bibitem{Poizat} \textsc{B.~Poizat}, \textsl{A Course in Model Theory: An Introduction to Contemporary Mathematical Logic}, Universitext (Springer, New York, 2000), \href{https://doi.org/10.1007/978-1-4419-8622-1}{doi:10.1007/978-1-4419-8622-1}.
		
		\bibitem{Pollard} \textsc{D.~Pollard}, \textsl{Convergence of Stochastic Processes}, Springer Ser.\ Stat.\ (Springer, New York, 1984), \href{https://doi.org/10.1007/978-1-4612-5254-2}{doi:10.1007/978-1-4612-5254-2}.
		
		\bibitem{Sauer} \textsc{N.~Sauer}, `On the Density of Families of Sets' \textsl{J.\ Comb.\ Theory, Ser.~A} \textbf{13} (1972) 145--147, \href{https://doi.org/10.1016/0097-3165(72)90019-2}{doi:10.1016/0097-3165(72)90019-2}.
		
		\bibitem{UnderstandingML} \textsc{S.~Shalev-Shwartz} and \textsc{S.~Ben-David}, \textsl{Understanding Machine Learning: From Theory to Algorithms} (Cambridge University Press, New York, 2014), \href{https://doi.org/10.1017/CBO9781107298019}{doi:10.1017/CBO9781107298019}.
		
		\bibitem{Shelah1971} \textsc{S.~Shelah}, `Stability, the f.c.p., and Superstability; Model Theoretic Properties of Formulas in First Order Theory', \textsl{Ann.\ Math.\ Logic} \textbf{3} (1971) 271--362, \href{https://doi.org/10.1016/0003-4843(71)90015-5}{doi:10.1016/0003-4843(71)90015-5}.
		
		\bibitem{Shelah1972} \textsc{S.~Shelah}, `A Combinatorial Problem; Stability and Order for Models and Theories in Infinitary Languages', \textsl{Pac.\ J.\ Math.\ }\textbf{41} (1972) 247--261, \href{https://doi.org/10.2140/pjm.1972.41.247}{doi:10.2140/pjm.1972.41.247}.
		
		\bibitem{Simon} \textsc{P.~Simon}, \textsl{A Guide to NIP Theories}, Lect.\ Notes Log.\ 44 (Association for Symbolic Logic, Cambridge University Press, Cambridge, 2015), \href{https://doi.org/10.1017/CBO9781107415133}{doi:10.1017/CBO9781107415133}.
		
		\bibitem{Steinhorn} \textsc{C.~Steinhorn}, `A brief introduction to o-minimality', \textsl{O-minimal Structures, Lisbon 2003: Proceedings of a Summer School by the European Research and Training Network RAAG} (Eds.\ M.~Edmundo, D.~Richardson and A.~J.~Wilkie), Lect.\ Notes Real Algebr.\ Anal.\ Geom.\ (Cuvillier, Göttingen, 2005) 11--31. 
		
		\bibitem{Valiant} \textsc{L.~G.~Valiant}, `A Theory of the Learnable', \textsl{Commun.\ ACM} \textbf{27} (1984) 1134--1142, \href{https://doi.org/10.1145/1968.1972}{doi:10.1145/1968.1972}.
		
		\bibitem{VC1968} \textsc{V.~N.~Vapnik} and \textsc{A.~Ja.~\v{C}ervonenkis}, `Uniform Convergence of Frequencies of Occurrence of Events to Their Probabilities', \textsl{Dokl.\ Akad.\ Nauk SSSR} \textbf{181} (1968) 781--783 (Russian), \textsl{Sov.\ Math., Dokl.\ }\textbf{9} (1968) 915--918 (English). 
		
		\bibitem{VC1971} \textsc{V.~N.~Vapnik} and \textsc{A.~Ya.~Chervonenkis}, `On the Uniform Convergence of Relative Frequencies of Events to Their Probabilities', \textsl{Teor.\ Veroyatn.\ Primen.\ }\textbf{16} (1971) 264--279 (Russian), \textsl{Theory Probab.\ Appl.\ }\textbf{16} (1971) 264--280 (English), \href{https://doi.org/10.1137/1116025}{doi:10.1137/1116025}. 
		
		\bibitem{Vidyasagar} \textsc{M.~Vidyasagar}, \textsl{Learning and Generalisation: With Applications to Neural Networks}, 2nd edn., Commun.\ Control Eng.\ (Springer, London, 2003), \href{https://doi.org/10.1007/978-1-4471-3748-1}{doi:10.1007/978-1-4471-3748-1}.
		
		\bibitem{Wilkie}  
		\textsc{A.~J.~Wilkie}, 
		`Model completeness results for expansions of the ordered field of real numbers by restricted Pfaffian functions and the exponential function', 
		\textsl{J.\ Amer.\ Math.\ Soc.} 9 (1996) 1051--1094, \href{https://doi.org/10.1090/S0894-0347-96-00216-0}{doi:10.1090/S0894-0347-96-00216-0}.
		
	\end{thebibliography}
\end{document}